\documentclass[10pt,twocolumn,letterpaper]{article}

\usepackage[pagenumbers]{cvpr} 

\usepackage[dvipsnames]{xcolor}

\definecolor{cvprblue}{rgb}{0.21,0.49,0.74}
\usepackage[pagebackref,breaklinks,colorlinks,allcolors=cvprblue]{hyperref}

\usepackage[ruled]{algorithm2e}
\usepackage[export]{adjustbox}
\usepackage{threeparttable}
\usepackage{subcaption}
\usepackage{multirow}
\usepackage{enumitem}
\usepackage{newfloat}
\usepackage{listings}
\usepackage{colortbl}
\usepackage{arydshln}
\usepackage{amsfonts}
\usepackage{caption}
\usepackage{wrapfig}
\usepackage{amssymb}
\usepackage{pifont}
\usepackage{float}
\usepackage{bbm}
\usepackage{bm}
\usepackage{mathrsfs}  

\definecolor{mygray}{gray}{.9}
\definecolor{mygreen}{RGB}{93,173,85}
\definecolor{mywarning}{RGB}{233,144,61}

\definecolor{DarkBlue}{RGB}{64,101,149}
\definecolor{azure}{rgb}{0.0, 0.5, 1.0}
\definecolor{gray}{rgb}{0.3, 0.3, 0.3}
\definecolor{DarkGreen}{RGB}{42,110,63}
\definecolor{DarkYellow}{RGB}{191,144,0}
\definecolor{DarkRed}{rgb}{0.6, 0, 0}

\newcolumntype{x}[1]{>{\centering\arraybackslash}p{#1pt}}
\newcolumntype{I}{!{\vrule width 1pt}}

\usepackage{tcolorbox}
\tcbuselibrary{theorems}
\definecolor{lightgray}{gray}{.9}
\definecolor{deepgray}{gray}{.8}
\tcbset{highlight math/.append style={left=0mm,right=0mm,top=0mm,bottom=0mm, colframe=white}}

\makeatletter
\newcommand{\thickhline}{%
    \noalign {\ifnum 0=`}\fi \hrule height 1pt
    \futurelet \reserved@a \@xhline
}

\makeatletter
\newenvironment{fullitemize}
{
\vspace{-1pt}
\begin{itemize}[leftmargin=*]
\setlength{\itemsep}{5pt}
\setlength{\parsep}{-5pt}
\setlength{\parskip}{-3pt}
\setlength{\leftmargin}{-10pt}
}
{
\end{itemize}
\vspace{-1pt}
}
\makeatother

\newcommand{\myhyperlink}[3][black]{\hyperlink{#2}{\color{#1}{#3}}}

\usepackage[capitalize]{cleveref}
\crefname{proposition}{Prop.}{Props.}
\crefname{section}{Sec.}{Secs.}
\crefname{table}{Tab.}{Tabs.}
\newcommand{\pub}[1]{{\color{gray}{\footnotesize{[{#1}]}}}}

\usepackage{xspace}
\makeatletter
\DeclareRobustCommand\onedot{\futurelet\@let@token\@onedot}
\def\@onedot{\ifx\@let@token.\else.\null\fi\xspace}
\def\eg{\textit{e.g}\onedot} 
\def\ie{\textit{i.e}\onedot}

\definecolor{darksalmon}{rgb}{0.91, 0.59, 0.48}
\definecolor{emerald}{rgb}{0.31, 0.78, 0.47}
\definecolor{green(pigment)}{rgb}{0.0, 0.65, 0.31}
\definecolor{amaranth}{rgb}{0.9, 0.17, 0.31}
\definecolor{iris}{rgb}{0.35, 0.31, 0.81}
\definecolor{uu}{rgb}{0.95, 0.51, 0.51}
\definecolor{spirodiscoball}{rgb}{0.06, 0.75, 0.99}
\usepackage{bbding}

\def\paperID{2563} 
\def\confName{CVPR}
\def\confYear{2025}

\title{Learn from Downstream and Be Yourself in \\ Multimodal Large Language Model Fine-Tuning}

\author{
Wenke Huang$^{1}$, \hspace{1pt} 
Jian Liang$^{1}$, \hspace{1pt} 
Zekun Shi$^{1}$, \hspace{1pt}
Didi Zhu$^{2}$, \hspace{1pt} \\ 
Guancheng Wan$^{1}$, \hspace{1pt} 
He Li$^{1}$, \hspace{1pt} 
Bo Du$^{1}$, \hspace{1pt}
Dacheng Tao$^{3}$, \hspace{1pt}
Mang Ye$^{1}$
\\ $^{1}$ Wuhan University. $^{2}$ Zhejiang University. $^{3}$ Nanyang Technological University\\
}

\begin{document}
\maketitle

\begin{abstract}
Multimodal Large Language Model (MLLM) have demonstrated strong generalization capabilities across diverse distributions and tasks, largely due to extensive pre-training datasets. Fine-tuning MLLM has become a common practice to improve performance on specific downstream tasks. However, during fine-tuning, MLLM often faces the risk of forgetting knowledge acquired during pre-training, which can result in a decline in generalization abilities. To balance the trade-off between generalization and specialization, we propose measuring the parameter importance for both pre-trained and fine-tuning distributions, based on frozen pre-trained weight magnitude and accumulated fine-tuning gradient values. We further apply an importance-aware weight allocation strategy, selectively updating relatively important parameters for downstream tasks. We conduct empirical evaluations on both image captioning and visual question-answering tasks using various MLLM architectures. The comprehensive experimental analysis demonstrates the effectiveness of the proposed solution, highlighting the efficiency of the crucial modules in enhancing downstream specialization performance while mitigating generalization degradation in  MLLM Fine-Tuning.
\end{abstract}

\newcommand{\attention}{{Attention}}
\newcommand{\bert}{{BERT}}
\newcommand{\vit}{{ViT}}
\newcommand{\cpvt}{{CPVT}}
\newcommand{\deit}{{DeiT}}
\newcommand{\convit}{{ConViT}}
\newcommand{\ceit}{{CeiT}}
\newcommand{\localvit}{{LocalViT}}
\newcommand{\swin}{{Swin}}
\newcommand{\cvt}{{CvT}}
\newcommand{\cswin}{{CSWin}}
\newcommand{\llava}{{LLaVA}}
\newcommand{\tailor}{{Tailor}}
\newcommand{\bioclip}{{BioCLIP}}
\newcommand{\tcprompt}{{TCP}}
\newcommand{\tac}{{TAC}}
\newcommand{\ipt}{{IPT}}
\newcommand{\vila}{{VILA}}
\newcommand{\grafting}{{Grafting}}
\newcommand{\gps}{{GPS}}
\newcommand{\argue}{{ArGue}}
\newcommand{\dora}{{DoRA}}
\newcommand{\blip}{{BLIP}}
\newcommand{\bliptwo}{{BLIP-2}}
\newcommand{\qwenvl}{{QWen-VL}}
\newcommand{\llavahr}{{LLaVA-HR}}
\newcommand{\palm}{{PaLM}}
\newcommand{\smop}{{SMoP}}
\newcommand{\mpvit}{{MPViT}}
\newcommand{\cropa}{{CroPA}}
\newcommand{\spu}{{SPU}}
\newcommand{\fullft}{{Full Fine-Tuning}}
\newcommand{\fullftabbr}{{Full FT}}
\newcommand{\halfft}{{Half Fine-Tuning}}
\newcommand{\halfftabbr}{{Half FT}}
\newcommand{\lmbff}{{LM-BFF}}
\newcommand{\llama}{{LLaMA}}
\newcommand{\vicuna}{{Vicuna}}
\newcommand{\cpr}{{CPR}}
\newcommand{\spt}{{SPT}}
\newcommand{\flora}{{Flora}}
\newcommand{\coprompt}{{CoPrompt}}
\newcommand{\dare}{{DARE}}
\newcommand{\lth}{{LTH}}
\newcommand{\tasl}{{TaSL}}
\newcommand{\fish}{{FISH}}
\newcommand{\wanda}{{Wanda}}
\newcommand{\rlcf}{{RLCF}}
\newcommand{\stprompt}{{STP}}
\newcommand{\neglabel}{{NegLabel}}
\newcommand{\restore}{{RESTORE}}
\newcommand{\prompttuning}{{Prompt Tuning}}
\newcommand{\dpl}{{DPL}}
\newcommand{\ptp}{{PTP}}
\newcommand{\swarm}{{SWARM}}
\newcommand{\damp}{{DAMP}}
\newcommand{\dpr}{{DPR}}
\newcommand{\saco}{{SaCo}}
\newcommand{\cpt}{{CPT}}
\newcommand{\cmpa}{{CMPA}}
\newcommand{\vars}{{VARS}}
\newcommand{\krona}{{Krona}}
\newcommand{\coda}{{CoDA}}
\newcommand{\clipvip}{{CLIP-ViP}}
\newcommand{\vipllava}{{ViP-LLaVA}}
\newcommand{\ciat}{{CIAT}}
\newcommand{\sift}{{SIFT}}
\newcommand{\dits}{{DiTs}}
\newcommand{\aprompt}{{Aprompt}}
\newcommand{\absvit}{{AbSViT}}
\newcommand{\vlp}{{Vision-Language Pre-trained Models}}
\newcommand{\vlpab}{{VLPM}}
\newcommand{\san}{{SAN}}
\newcommand{\clapaudio}{{CLAP}}
\newcommand{\cascadeclip}{{Cascade-CLIP}}
\newcommand{\declip}{{DeCLIP}}
\newcommand{\imagebind}{{ImageBind}}
\newcommand{\uavod}{{UAV-OD}}
\newcommand{\clip}{{CLIP}}
\newcommand{\bop}{{BoP}}
\newcommand{\bicro}{{BiCro}}
\newcommand{\defo}{{DeFO}}
\newcommand{\flamingo}{{Flamingo}}
\newcommand{\ssam}{{SSAM}}
\newcommand{\alip}{{ALIP}}
\newcommand{\prefixtuning}{{PrefixTuning}}
\newcommand{\npt}{{NPT}}
\newcommand{\dac}{{DAC}}
\newcommand{\fsam}{{FSAM}}
\newcommand{\da}{}
\newcommand{\livt}{{LiVT}}
\newcommand{\instructblip}{{InstructBLIP}}
\newcommand{\stwo}{{S$^{2}$}}
\newcommand{\alignVL}{{ALIGN}}
\newcommand{\ramt}{{R-AMT}}
\newcommand{\adamvmoe}{{AdaMVMoE}}
\newcommand{\tpt}{{TPT}}
\newcommand{\graphadapter}{{GraphAdapter}}
\newcommand{\lingualsmoe}{{Lingual-SMoE}}
\newcommand{\tai}{{TaI}}
\newcommand{\dapt}{{DAPT}}
\newcommand{\dualprompt}{{DualPrompt}}
\newcommand{\dept}{{DePT}}
\newcommand{\logsparse}{{LogSparse}}
\newcommand{\fuller}{{FULLER}}
\newcommand{\depthclip}{{DepthCLIP}}
\newcommand{\vlab}{{VLAB}}
\newcommand{\lpt}{{LPT}}
\newcommand{\tvp}{{TVP}}
\newcommand{\promptkd}{{PromptKD}}
\newcommand{\metatrans}{{Meta-Transformer}}
\newcommand{\gasam}{{GA-SAM}}
\newcommand{\sprompt}{{SP}}
\newcommand{\promptstyler}{{PromptStyler}}
\newcommand{\umt}{{UMT}}
\newcommand{\plot}{{PLOT}}
\newcommand{\doprompt}{{DoPrompt}}
\newcommand{\misa}{{MISA}}
\newcommand{\protext}{{ProText}}
\newcommand{\vipt}{{ViPT}}
\newcommand{\kgcoop}{{KgCoOp}}
\newcommand{\bridge}{{Birdge}}
\newcommand{\dferc}{{DFERC}}
\newcommand{\xprompt}{{Xprompt}}
\newcommand{\dualcoopplus}{{DualCoOP++}}
\newcommand{\upt}{{UPT}}
\newcommand{\itwomcl}{{I$^2$MCL}}
\newcommand{\gblending}{{G-Blending}}
\newcommand{\greedy}{{Greedy}}
\newcommand{\mmanet}{{MMANet}}
\newcommand{\grda}{{GRDA}}
\newcommand{\lmf}{{LMF}}
\newcommand{\kapt}{{KAPT}}
\newcommand{\mfm}{{MFM}}
\newcommand{\etwovpt}{{E$^2$VPT}}
\newcommand{\audioclip}{{AudioCLIP}}
\newcommand{\hotprotein}{{HotProtein}}
\newcommand{\prac}{{PRAC}}
\newcommand{\promptsrc}{{PromptSRC}}
\newcommand{\diffpurning}{{DiffPurning}}
\newcommand{\prograd}{{ProGrad}}
\newcommand{\bike}{{BIKE}}
\newcommand{\maple}{{MaPLe}}
\newcommand{\fdmer}{{FDMER}}
\newcommand{\mtwopt}{{M2PT}}
\newcommand{\film}{{FiLM}}
\newcommand{\laclip}{{LaCLIP}}
\newcommand{\selfmm}{{Self-MM}}
\newcommand{\conc}{{Concatenation}}
\newcommand{\summ}{{Summation}}
\newcommand{\clipreid}{{CLIP-REID}}
\newcommand{\gated}{{Gated}}
\newcommand{\capforvideo}{{Cap4Video}}
\newcommand{\agm}{{AGM}}
\newcommand{\icode}{{i-Code}}
\newcommand{\msrl}{{MSRL}}
\newcommand{\mslr}{{MSLR}}
\newcommand{\fagm}{{FAGM}}
\newcommand{\dualcoop}{{DualCoOp}}
\newcommand{\saf}{{SAF}}
\newcommand{\shape}{{SHAPE}}
\newcommand{\rigorllm}{{RigorLLM}}
\newcommand{\aptm}{{APTM}}
\newcommand{\fishr}{{Fishr}}
\newcommand{\soho}{{SOHO}}
\newcommand{\ijepa}{{I-JEPA}}
\newcommand{\pmr}{{PMR}}
\newcommand{\smil}{{SMIL}}
\newcommand{\man}{{MAN}}
\newcommand{\mbt}{{MBT}}
\newcommand{\dmd}{{DMD}}
\newcommand{\pmf}{{PMF}}
\newcommand{\tipadater}{{Tip-Adapter}}
\newcommand{\clipadapter}{{CLIP-Adapter}}
\newcommand{\coop}{{CoOP}}
\newcommand{\softmask}{{SoftMask}}
\newcommand{\clippo}{{CLIPPO}}
\newcommand{\cocoop}{{CoCoOp}}
\newcommand{\cafo}{{CaFo}}
\newcommand{\pcme}{{PCME}}
\newcommand{\ape}{{APE}}
\newcommand{\ogmge}{{OGMGE}}
\newcommand{\vificlip}{{ViFi-CLIP}}
\newcommand{\fdt}{{FDT}}
\newcommand{\shaspec}{{ShaSpec}}
\newcommand{\cleanclip}{{CleanCLIP}}
\newcommand{\fourierft}{{FourierFT}}
\newcommand{\vilt}{{ViLT}}
\newcommand{\metaadapter}{{MetaAdapter}}
\newcommand{\gmc}{{GMC}}
\newcommand{\denseclip}{{DenseCLIP}}
\newcommand{\chils}{{CHiLS}}
\newcommand{\mtwofnet}{{M2FNet}}
\newcommand{\mmcosine}{{MMCosine}}
\newcommand{\lora}{{LoRA}}
\newcommand{\offsitetuning}{{Offsite-Tuning}}
\newcommand{\adapter}{{Adapters}}
\newcommand{\pst}{{PST}}
\newcommand{\lst}{{LST}}
\newcommand{\ltsft}{{LT-SFT}}
\newcommand{\moefication}{{MoEfication}}
\newcommand{\toast}{{TOAST}}
\newcommand{\adalora}{{AdaLoRA}}
\newcommand{\pafi}{{PaFi}}
\newcommand{\vpt}{{VPT}}
\newcommand{\infoprompt}{{InfoPrompt}}

\newcommand{\lonereg}{{L1-Regularization}}
\newcommand{\ltworeg}{{L2-Regularization}}
\newcommand{\ltworegabbrv}{{L2-Reg}}

\newcommand{\gqa}{{GQA}}
\newcommand{\textvqa}{{TextVQA}}
\newcommand{\ocrvqa}{{OCRVQA}}
\newcommand{\okvqa}{{OKVQA}}
\newcommand{\vizwiz}{{Vizwiz}}
\newcommand{\scienceqa}{{ScienceQA}}
\newcommand{\flickrthirtyk}{{Flickr30k}}
\newcommand{\iconqa}{{IconQA}}
\newcommand{\vqavtwo}{{VQAV2}}
\newcommand{\cococap}{{COCO-Capation}}

\newcommand{\llm}{{Large Language Models}}
\newcommand{\llmabbrv}{{LLM}}
\newcommand{\mllm}{{Multimodal Large Language Model}}
\newcommand{\mllmabbrv}{{MLLM}}

\newcommand{\ours}{{Specialization via Importance Discrepancy Evaluation for Refinement}}
\newcommand{\oursabbrv}{{SPIDER}}

\newcommand{\first}{{Importance Discrepancy Measurement}}
\newcommand{\firstabbrv}{{IDM}}

\newcommand{\second}{{Importance Selection Mask}}
\newcommand{\secondabbrv}{{ISM}}

\newcommand{\zeroshot}{{Zero-shot}}

\newcommand{\bluedown}[1]{$_{\color{BlueGreen}\downarrow #1}$}
\newcommand{\redup}[1]{$_{\color{RedOrange}\uparrow #1}$}

\section{Introduction}
Recent years have witnessed remarkable progress in \mllm{} (\mllmabbrv{}),  which have demonstrated impressive competency in various vision-understanding tasks {\cite{LLaVA_NeurIPS23,InstructBLIP_NeurIPS23,VILA_CVPR24,LLaVA15_CVPR24}}. \mllmabbrv{} generally follows the paradigm to fuse the pre-trained vision encoder {\cite{CLIP_ICML21,ViT_ICLR21}} into the representation space of the  \llm{} (\llmabbrv), \eg, \llama{} {\cite{LLaMA_arXiv23}} and \vicuna{} {\cite{Vicuna_arXiv23}}, via the connector module {\cite{InstructBLIP_NeurIPS23,LLaVA_NeurIPS23,LLaVAHR_arXiv24}}. Considering that \mllm{} is optimized on huge-scale and various-type multimodality instruction-following datasets {\cite{COCO_ECCV14,TextVQA_CVPR19,OCRVQA_ICDAR19}}, it brings powerful generalization ability on different related tasks. Despite this, \mllmabbrv{} still performs poorly on downstream datasets {\cite{SpeandGenonCF_arXiv23,CFforLLMinCFT_arXiv23,CFinMLLMFT_CPAL24,UnifiedDelta_arXiv24,TwinMerging_NeurIPS24}}. 

\begin{figure}[t]
\centering
\includegraphics[width=\linewidth]{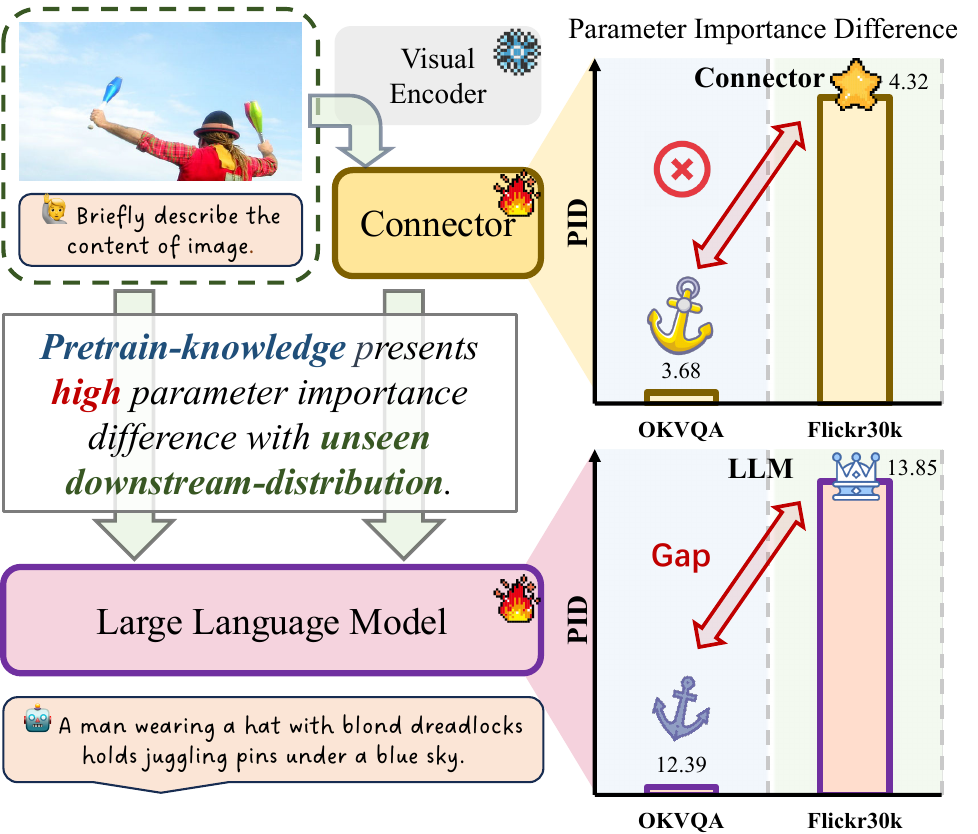}
\vspace{-15pt}
\captionsetup{font=small}
\caption{\textbf{Background and Motivation}. Fine-tuning \mllm{} (\mllmabbrv{}) on downstream tasks typically involves training (\protect\includegraphics[scale=0.4,valign=c]{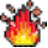}) the connector and \llmabbrv{} modules, and freezing (\protect\includegraphics[scale=0.4,valign=c]{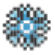}) the visual encoder. We reveal a {\color{DarkRed}\textbf{higher}} parameter importance difference (\textbf{PID}) on {\color{DarkGreen}\textbf{unseen downstream distributions}}, \eg, \flickrthirtyk{}, compared to {\color{DarkBlue}\textbf{seen upstream distribution}}, \eg, \okvqa{}. PID $= cos(|w^*|,|g|)^{-2}$. We utilize the absolute value of the pre-trained weight $|w^*|$  and and fine-tuning gradients $|g|$ to represent the upstream and downstream parameter importance.} 
\label{fig:problem}
\vspace{-10pt}
\end{figure}

The common practice is to fine-tune foundation models on specific tasks to enhance task performance or align the model behavior with human expectations {\cite{PEFTMLLM_ACL24,PEFTSurvey_arXiv24}}. Specifically, existing solutions normally freeze the visual encoder, focusing solely on connector layers and the LLM component {\cite{PandaGPT_arXiv23,Honeybee_CVPR24,SPHINX_ECCV24}}. Thus, during the fine-tuning stage, the \mllmabbrv{} gains specialization ability to achieve exceptional performance on the fine-tuning task. However, since the small fine-tuning dataset does not have sufficient coverage of the distribution as well as tasks, the fine-tuned model can potentially lose its generality which is acquired through pre-training stage. The effect of deteriorating the model previous generic knowledge upon new learning is a well-documented challenge, referred to as catastrophic forgetting {\cite{Connectionist_PR90,CataInter_PLM89,Catastrophic_99,CFforLLMinCFT_arXiv23,CFinMLLMFT_CPAL24}}. Consequently, a pivotal question raises: \textit{How to enhance the specialization ability on target tasks while maintaining generalization knowledge for \mllm{} Fine-Tuning?}

In our work, we propose a simple yet effective method, \textbf{Sp}ecialization via \textbf{I}mportance \textbf{D}iscrepancy \textbf{E}valuation for \textbf{R}efinement, abbreviated as \oursabbrv{} (\protect\includegraphics[scale=0.1,valign=c]{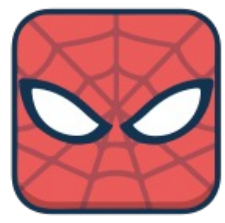}) to simultaneously accommodate task-oriented properties and maintain task-generic knowledge to alleviate the catastrophic forgetting phenomenon. Own to the over-parameterized characteristics of the deep neural network \cite{CalibrationDNN_ICML17,UncertainEstimateDeepEnsembles_NeurIPS17}, we notice that {{not all parameters contribute equally to fit the target distribution}}, which has confirmed soundness in the relative researches {\cite{Pruning_ICLR17,LTH_ICLR19,RethinkingNetPruning_ICLR19,FISH_NeurIPS21,Chase_NeurIPS23,Wanda_ICLR24}}. Consequently, we assert that pre-training and fine-tuning distributions exhibit distinct parameter importance degrees for representing generic and specialized knowledge. In line with this, we are motivated to \textit{\textbf{selectively update relatively important parameters for the downstream task while preserving the remaining for both generalization and specialization ability}}.

Driven by this motivation, the above problem becomes more fundamental: \hypertarget{Q1}{\textbf{\uppercase\expandafter{\romannumeral1})}}: 
\textit{How to measure parameter importance for generic and specialized knowledge}? \hypertarget{Q2}{\textbf{\uppercase\expandafter{\romannumeral2})}}:
\textit{How to selectively maintain and optimize during downstream learning}.
To respond to question \myhyperlink{Q1}{\textbf{\uppercase\expandafter{\romannumeral1})}}, we introduce the \first{} (\firstabbrv{}) to quantify the parameter importance degree towards the generalized and specialized knowledge. With respect to the prior generic information, we take inspiration from {\cite{MagnitudePurning_NeurIPS15,LTH_ICLR19,Wanda_ICLR24}} and argue that \textit{weight magnitude} positive correlates with the prediction tendency. Thus, we adopt the  \mllmabbrv{} \textbf{pretrained weight magnitude} to measure the parameter importance towards the generalization attitude. Towards the posterior specialized behavior, we utilize \textbf{current gradient norm} to pinpoint regions that learn crucial downstream knowledge. The rationale behind this is that \textit{gradients trajectories} directly provides intensity information of the learning signal imposed on each parameter element for optimization objective {\cite{Fisher_1922,NatureGradient_arXiv13,EWC_PNAS17,Snip_ICLR19,StableCL_NeurIPS20,MovementPruning_NeurIPS20,SIFT_ICML24}}.  By comparing the parameter importance rank, we can differentiate between generic and task-specific parameters, As illustrated in \cref{fig:problem}. tuning on the unseen fine-tuning datasets, \eg, \flickrthirtyk{}, exhibits a more pronounced parameter importance discrepancy than the seen \okvqa{} distributions. This observation further supports our motivation to mitigate catastrophic forgetting in \mllmabbrv{} by considering parameter importance discrepancy. Driven by question \myhyperlink{Q2}{\textbf{\uppercase\expandafter{\romannumeral2})}}, we then propose the \second{} (\secondabbrv{}) to selectively consolidate or optimize candidate parameters for the target distribution. Specifically, during the network backward pass, we identify and consolidate parameters that exhibit relatively higher importance for general task knowledge, while optimizing the remaining elements to enhance task-specific performance. For thorough examination, we conduct experiments on two representative \mllmabbrv{}: \vila{} {\cite{VILA_CVPR24}} and \llava{} {\cite{LLaVA_NeurIPS23,LLaVA15_CVPR24}}. We fine-tune on two major tasks image-captioning and visual question answering and evaluate the generic knowledge on the pre-trained seen datasets {\cite{VQAV2_CVPR17,OKVQA_CVPR19,GQA_CVPR19,TextVQA_CVPR19}}.  The main contributions are summarized as follows:

\begin{fullitemize}
\item We focus on addressing the catastrophic forgetting problem in fine-tuning \mllm{} (\mllmabbrv{}) for downstream tasks. We reveal that due to the distribution shift between upstream and downstream patterns, parameters exhibit varying degrees of importance.

\item We introduce \oursabbrv{}, which measures generalization and specialization based on the behavior of frozen weights and updating gradients. Our method identifies relatively important elements for the downstream task and conduct critical-aware weight allocation on candidate parameters. By selecting and ranking these elements, our approach offers a novel solution to effectively tackle the generalization and specialization dilemma in \mllmabbrv{}.

\item We conduct a comprehensive analysis on four downstream datasets: \flickrthirtyk{} {\cite{Flickr_TACL14}}, \cococap{} {\cite{COCO_ECCV14}}, \scienceqa{} {\cite{ScienceQA_NeurIPS22}}, and \iconqa{} {\cite{IconQA_NeurIPS21}}, using \vila{} {\cite{VILA_CVPR24}} and \llava{} {\cite{LLaVA_NeurIPS23}} architectures. Along with a series of ablation studies, the promising results empirically validate the effectiveness of \oursabbrv{} in improving fine-tuning performance and mitigating generalization forgetting.
\end{fullitemize}

\section{Related Works}
\subsection{Multimodal Large Language Models}  With the impressive success of \llm{} (\llmabbrv), such as GPT {\cite{GPT2_OPENAI19,GPT3_NeurIPS20,GPT4_arXiv23}}, \llama{} \cite{LLaMA_arXiv23}, \vicuna{} {\cite{Vicuna_arXiv23}}, \palm{} \cite{PaLM_arXiv22,PaLMv2_arXiv23}, growing interest has been aroused in building end-to-end \mllm{} (\mllmabbrv{}), \eg, \flamingo{} \cite{Flamingo_NeurIPS22}, \bliptwo{} \cite{BLIP_ICML22,BLIPv2_ICML23}, \instructblip{} \cite{InstructBLIP_NeurIPS23}, \qwenvl{} {\cite{QwenVL_arXiv23}}, \llava{} \cite{LLaVA_NeurIPS23,LLaVA15_CVPR24,LLaVA-Phi_arXiv24}, \vila{} \cite{VILA_CVPR24}. Existing \mllmabbrv{} solutions normally follow to utilize the visual extractor {\cite{CLIP_ICML21,ViT_ICLR21}} to encode visual features and utilize the connector module to project visual tokens into word embedding space of the \llmabbrv{}, \ie, treating visual input as the foreign language {\cite{BEiT3_CVPR23}}. Then, the visual and textual tokens are concatenated and fed into the \llmabbrv{}. The \llmabbrv{} is used to accomplish various vision-language tasks in an auto-regressive manner. For example, the famous \mllmabbrv{} work, \llava{} \cite{LLaVA_NeurIPS23} adopts a linear projection layer to connect the visual encoder and the \llmabbrv{} {\cite{Vicuna_arXiv23,LLaMA_arXiv23}}. Despite their effectiveness, existing works primarily emphasize the generalization ability across various tasks, resulting in the constrained performance on specific downstream target tasks. Therefore, it is an intuitive solution to fine-tune the \mllmabbrv{} in order to enhance the performance on the particular task.  

\subsection{Catastrophic Forgetting in \mllm{} Fine-Tuning}
\label{sec:FTandCFforMLLM}
Commonly optimized on downstream tasks {\cite{CE_AOR05}}, deep neural network is empirically proved to suffer from the \textit{catastrophic forgetting} problem {\cite{Connectionist_PR90,CataInter_PLM89,Catastrophic_99,CFforLLMinCFT_arXiv23,EWC_PNAS17,CFinMLLMFT_CPAL24}}, a significant issue where models forget previously learned information when exposed to new data. In the context of \mllmabbrv{}, this results in catastrophic forgetting of generic knowledge, which severely impairs the model transferability across previously learned datasets. Therefore, balancing the ability to fit downstream tasks while maintaining generalization becomes a crucial challenge for \mllm{}.
Existing methods could be roughly divided into four categories {\cite{SpeandGenonCF_arXiv23,PEFTSurvey_arXiv24,PEFTforVisionSurvey_arXiv24}}. 
\textbf{i)} \textit{Additive Parameter} Learning {\cite{Adapter_ICML19,LST_NeurIPS22,PromptTuning_arXiv21,LPT_arXiv22,CoOp_IJCV22,CoCoOp_CVPR22,TipAdapter_ECCV22}} primarily focus on strategically incorporating additional trainable parameters within the architecture. For example, adapter {\cite{Adapter_ICML19,TipAdapter_ECCV22,GraphAdapter_NeurIPS23,ClipAdapter_IJCV23,PormptAdapter_arXiv23}} typically consist of multi-layer perceptions and residual connections {\cite{ResNet_CVPR16}} that combine pre-trained features with updated ones.  Additionally, prompt {\cite{PromptTuning_arXiv21,CoOp_IJCV22,CoCoOp_CVPR22,UPT_arXiv22,TPT_NeurIPS22,DualPrompt_ECCV22}} directly appends adjustable vectors to the input sequence. 
\textbf{ii)} \textit{Reparameterization Tuning} {\cite{LoRA_ICLR22,AdaLoRA_arXiv23,LiGO_ICLR23,Flora_ICML24,DoRA_ICML24}} also introduce new learnable parameters during the training stage, which are then integrated into the original \mllmabbrv{} through reparameterization during inference. For instance, \lora{} \cite{LoRA_ICLR22} assumes that the changes in linear model weights follow a low-rank behavior. Despite the certain advantages, these two research streams introduce additional parameters into the pre-trained model and disrupt the original architecture, leading to increased computational costs and presents restricted architecture compatibility. 
\textbf{iii)} \textit{Regularization-based Optimization} {\cite{EWC_PNAS17,SynapticCL_ICML17,InductiveBias_ICML18,StructuredLaplace_NeurIPS18,GCL_NeurIPS20,FedProx_MLSys2020,Grafting_ICML23}} introduce the loss constraints to preserve the previously learnt knowledge. Several studies add regularization terms to the loss functions to penalize parameter changes and mitigate catastrophic forgetting. However, aforementioned solutions require to modify the loss function and thus conflict with personalized fine-tuning loss design.  
\textbf{iv)} \textit{Partial-based Updating} \cite{PST_IJCAI22,LTSFT_ACL22,LoSparse_ICML23,DARE_ICML24,GPS_CVPR24,SPU_CVPR24,ModelTailor_ICML24,TwinMerging_NeurIPS24} focuses on modifying a subset of downstream-relevant parameters, making it architecture-agnostic and orthogonal to the downstream loss objective. For instance, \gps{} \cite{GPS_CVPR24} and \spu{} \cite{SPU_CVPR24} perform sparse updates based on gradient signals, while \dare{} \cite{DARE_ICML24} and \tailor{} \cite{ModelTailor_ICML24} operate on delta parameters. However, previous methods struggle to retain generic knowledge and their performance is highly sensitive to predefined selection thresholds. In our research, recognizing the distinct characteristics of deep neural networks, we argue that parameters exhibit differing importance distributions between pre-training and fine-tuning phases. Therefore, we measure parameter importance in a self-driven manner, selectively updating those with relatively higher importance for downstream tasks while preserving the generalization capability.

\begin{table}[t]\small
\centering
{
\resizebox{0.9\columnwidth}{!}{
\setlength\tabcolsep{3pt}
\renewcommand\arraystretch{1.1}
\begin{tabular}{cIc|c|c|c|c}
\hline\thickhline
\rowcolor{mygray}
Limitation & Add. & Repara. & Reg. & Part. &  Ours \\
\hline\hline
\textbf{Specify} Architecture 
& \textcolor{darksalmon}{\Checkmark} & \textcolor{darksalmon}{\Checkmark} & & &  \textcolor{green(pigment)}{\XSolidBrush} \\ 
\textbf{Modify} Optimization 
& & &  \textcolor{darksalmon}{\Checkmark} &  & \textcolor{green(pigment)}{\XSolidBrush} \\
\textbf{Require} Hyper-Parameter 
& & &  \textcolor{darksalmon}{\Checkmark} &  \textcolor{darksalmon}{\Checkmark} & \textcolor{green(pigment)}{\XSolidBrush}
\end{tabular}}}
\vspace{-10pt}
\captionsetup{font=small}
\caption{\textbf{Limitation} for different Anti-Forgetting \mllmabbrv{} methods: {Additive Parameter Learning} (Add.), Reparameterization Tuning (Repara.), Regularization based Optimization (Reg.), and Partial-based Updating (Part.). Refer to \cref{sec:FTandCFforMLLM} for details.
}
\label{tab:limitationcomparison}
\vspace{-20pt}
\end{table}


\section{Methodology}

\subsection{Preliminary}
\noindent Given the \mllm{} (\mllmabbrv{}) architecture, the \mllmabbrv{} model ($\theta$) typically includes three parts: visual encoder $f$, \eg, \vit{} \cite{ViT_ICLR21}, \llmabbrv{} ($g$), \eg, \vicuna{} {\cite{Vicuna_arXiv23}} and \llama{} {\cite{LLaMA_arXiv23}}, and the connector module $\varphi$ {\cite{LLaVA_NeurIPS23,InstructBLIP_NeurIPS23,LLaVA15_CVPR24,VILA_CVPR24}}. For a query instance, the input consists of both a visual image $x^v$ and a textual instruction $x^t$. The corresponding label is a language response $y$. First, we extract the visual features $z^v = f(x^v)$, and then apply the trainable projection $\varphi$ to convert $z^v$ into language embedding tokens, $h^v = \varphi \cdot z^v$. And textual token as $h^t = \text{Tokenize}(x^t)$. Next, we combine both visual and textual tokens and pass them into the \llmabbrv{} module $g$ to generate the language output $\hat{y} = g([h^v, h^t])$. In our work, following previous \mllmabbrv{} fine-tuning works and benchmarks {\cite{PEFTMLLM_ACL24,ModelTailor_ICML24}}, we select and fine-tune partial trainable parameter module $w$ from the \mllmabbrv{} model to adapt to the downstream task $\mathcal{T}$ with distribution ($D^\mathcal{T}$).  Normally, learnable modules are the connector module ($\varphi$) and candidate \llmabbrv{} ($g$) block layers as $w=\{\varphi,g\}$. This default \mllmabbrv{} optimization follows:
\begin{equation}\small
\setlength\abovedisplayskip{0pt} \setlength\belowdisplayskip{0pt}
\arg \min_{w} \mathbb{E}_{(x^v,x^t,y) \in {\mathcal{D}^\mathcal{T}}} \mathcal{L}\left(g([\varphi(h^v),h^t]), y \right).
\label{eq:ce_loss}
\end{equation}

\subsection{\ours{}}
To enhance downstream efficiency while preserving generic knowledge in \mllmabbrv{}, we assess parameter importance across pre-training and fine-tuning distributions, selectively updating downstream critical elements, including two components: \first{} (\firstabbrv{}  \cref{sec:first}) for ranking parameter importance, and \second{} (\secondabbrv{} \cref{sec:second}) for selective updates.

\subsubsection{\first{}}
\label{sec:first}
\noindent \textbf{Importance for Generalization Knowledge}. Generic knowledge embedded
in \mllmabbrv{} provides bases for strong performance in various domains and quick transfer to different tasks; when directly fine-tuning on newly received tasks with no regard to preserving its pre-existing, \mllmabbrv{} faces the catastrophic forgetting on the generalization ability. Thus, with respect to the generalization knowledge, we take inspiration from the magnitude pruning {\cite{MagnitudePurning_NeurIPS15}} and weight magnitude represents
how much the parameter contributes to the model prediction {\cite{LTH_ICLR19}}. Thus, in our work, we directly utilize the weight magnitude {\cite{MagnitudePurning_NeurIPS15,LTH_ICLR19,Wanda_ICLR24}} for pre-trained parameters $w^*$ to rank the generalization parameter importance $\mathcal{I}$ as the following formulation.
\begin{subequations}\small
\setlength\abovedisplayskip{0pt} \setlength\belowdisplayskip{0pt}
\begin{align}
\mathcal{I}[v] = |w^*[v]| & & \text{Absolute}, \label{eq:magabsolute} \\
\mathcal{I}[v] =\frac{\mathcal{I}[v]-\text{Mean}(\mathcal{I})}{\text{Std}(\mathcal{I})} & & \text{Normalization},
\label{eq:magnorm} \\ 
{\tcbhighmath[colback=SkyBlue!8]{\mathcal{I}[v]}} = \frac{1}{1 + e^{-\mathcal{I}[v]}} & & \text{Rescale}.
\label{eq:magrescale}
\end{align}
\end{subequations}
In this context, the notation $[v]$ represents the $n^{th}$ component value of a given tensor vector. The role of the above form is threefold. \cref{eq:magabsolute} computes the weight magnitude, and \cref{eq:magnorm} is applied to eliminate the effect of dimensional analysis. We further rescale into the bounded range for comparison via \cref{eq:magrescale}. Thus, $\mathcal{I}[v]$ is  within $(0,1)$

\noindent \textbf{Importance for Specialization Knowledge}.
With respect to fine-tuning the downstream task, we aim to identify which parameters are more relevant to the specific task at hand. We argue that the gradient signal acts an effective evaluation metric as the following formulation: 
\begin{equation}\small
\setlength\abovedisplayskip{0pt} \setlength\belowdisplayskip{0pt}
\delta [v] = \frac{\partial \mathcal{L}\left(g([\varphi(h^v),h^t]), y \right)}{\partial w[v]},
\label{eq:gradient}
\end{equation}
where $\delta [v]$ denotes the gradient of the loss function with respect to the parameter $w[v]$, evaluated at the query sample. The intuition is that parameters with larger gradient values correspond to directions where the loss function changes most rapidly, facilitating efficient gradient descent during fine-tuning. Thus, we derive the specialization parameter importance $\mathcal{G}$. The formulation is quantified as follows:
\begin{equation}\small
\setlength\abovedisplayskip{0pt} \setlength\belowdisplayskip{0pt}
\begin{split}
\label{eq:gradientweight}
\mathcal{G}[v] &= \text{Norm}(|\delta [v]|) \in (-\infty,\infty), \\
{\tcbhighmath[colback=green!8]{\mathcal{G}[v]}} &= \text{Sigmoid}(\mathcal{G}[v])) \in (0,1).
\end{split}
\end{equation}
Notably, due to the stochastic nature of sampling, $\delta [v]$ is unstable in revealing parameter importance, introducing significant uncertainty in estimating specialization knowledge. To mitigate this, we take inspiration from the momentum operation {\cite{MOCO_CVPR20,MOCOv2_arXiv20,MOCOv3_ICCV21}} and perform iterative accumulation of sample gradients to overcome this uncertainty. 

Therefore, we evaluate parameter importance for both generalization and specialization by utilizing pre-training weights and fine-tuning gradient information. To ensure a balanced assessment, we apply the consistent normalization and rescaling methods to these two metrics.
\subsubsection{\second{}}
\label{sec:second}
After localizing the parameter importance for both generalization ($\mathcal{I}$) and specialized knowledge ($\mathcal{G}$) during the fine-tuning stage, we only update the selected parameters while keeping the remaining pre-trained model parameters frozen. Thus, the straightforward approach is to treat the relatively important elements for the downstream task as candidate parameters for updates. Thus, we define the updating mask $\bm{M}$ as the following formulation:
\begin{equation}\small
\setlength\abovedisplayskip{0pt} \setlength\belowdisplayskip{0pt}
\label{eq:binarymask}
\bm{M}[v] = \left\{
\begin{array}{ll}
1, & \mathcal{G}[v] > \mathcal{I}[v], \\
0, &\text{else}.
\end{array}
\right.
\end{equation}
When $\bm{M}[v]=1$, the query parameter is selected as updating candidate. We denote the current  \mllmabbrv{} model as $w$.  We utilize the frozen pre-trained parameter  $w^*$  to reweight the current model, thereby restoring the original pre-trained knowledge conditions as follows:
\begin{equation}\small
\setlength\abovedisplayskip{0pt} \setlength\belowdisplayskip{0pt}
w = w \odot \bm{M} + w^* \odot (1-\bm{M}).
\label{eq:paramagg}
\end{equation}
However, this operation introduces no variance in the candidate parameter updates. Moreover, due to its normalization property, the aforementioned solution can be seen as masking fifty percent of the parameter updates, which still results in the degradation of generalization performance. We argue that for the selected parameters, assigning higher weights to those exhibiting a greater discrepancy in importance, and conversely lower weights to less significant parameters. Thus, we propose the \second{} (\secondabbrv{}) to reconstruct the aggregation weight in \cref{eq:binarymask} as:
\begin{equation}\small
\setlength\abovedisplayskip{1pt}\setlength\belowdisplayskip{1pt}
\label{eq:weightmask}
\bm{M}[v] = \left\{
\begin{array}{ll}
 \frac{\mathcal{G}[v]}{ \mathcal{G}[v]+ \mathcal{I}[v]}, &  \mathcal{G}[v] > \mathcal{I}[v], \\
0, & \text{else}.
\end{array}
\right.
\end{equation}
Furthermore, we rescale the aggregation weights based on the mean behavior of the selected elements, while restricting the upper bound to 1. This can be considered as the following rescale operation strategy:
\begin{equation}\small
\setlength\abovedisplayskip{1pt}\setlength\belowdisplayskip{1pt}
\label{eq:rescalemask}
{\tcbhighmath[colback=Goldenrod!15]{\bm{M}[v]}} = 
\begin{cases} 
\min (1, \frac{\bm{M}[v]}{\text{Mean}(\bm{M}[\bm{M} \neq 0])}), & {\tcbhighmath[colback=green!8]{\mathcal{G}[v]}} > {\tcbhighmath[colback=SkyBlue!8]{\mathcal{I}[v]}}, \\
0, & \text{else}.
\end{cases}
\end{equation}
Based on the above \second{} (\secondabbrv{}), we rewrite the \cref{eq:paramagg} to update the current model and plot the algorithm description in  \cref{alg:ours} and \cref{fig:ConceptCompar}.

\begin{algorithm}[t]
\caption{\oursabbrv{}}
\label{alg:ours}
\SetAlgoLined
\SetNoFillComment
\SetArgSty{textnormal}
\small{\KwIn{Fine-Tuning Epoch $E$, Overall \mllmabbrv{} Network $\theta$, Trainable parameter module $w$, Frozen Pre-trained parameter weight $w^*$, }}
\small{\KwOut{The optimized selected \mllmabbrv{} model $w$}}

\BlankLine
{\footnotesize{\color{DarkBlue}{\tcc{Generalization Knowledge Rank}}}}

$\mathcal{I} \leftarrow$ ($w^*$) in \cref{eq:magrescale}

\For {$e=1, 2, ..., E$}{

\For{$(x^v,x^t,y) \in {D}^k$}{

$h^v = \varphi(f(x^v)), h^t = \text{Tokenize}(x^t)$

$\delta = \nabla \mathcal{L}\left(g([h^v,h^t]), y \right)$ via \cref{eq:gradient}

{\footnotesize{\color{DarkBlue}{\tcc{Specialization Knowledge Rank}}}}

$\mathcal{G} \leftarrow$ ($\delta$) in \cref{eq:gradientweight}

{\footnotesize{\color{DarkBlue}{\tcc{Importance Selection Mask}}}}

$\bm{M}[v] = \left\{
\begin{array}{ll}
 \frac{\mathcal{G}[v]}{ \mathcal{G}[v]+ \mathcal{I}[v]}, & \mathcal{G}[v] > \mathcal{I}[v], \\
0, & \text{else}.
\end{array}
\right.$

$\downarrow$ \textbf{Rescale} Importance Mask Matrix $\bm{M}$

${\tcbhighmath[colback=Goldenrod!15]{\bm{M}[v]}} = 
\begin{cases} 
\min\left(1, \frac{\bm{M}[v]}{\text{Mean}(\bm{M}[\bm{M} \neq 0])}\right), & {\tcbhighmath[colback=green!8]{\mathcal{G}[v]}} > {\tcbhighmath[colback=SkyBlue!8]{\mathcal{I}[v]}}, \\
0, & \text{else}.
\end{cases}$

\BlankLine

$w  = w  - \eta \nabla \mathcal{L}$
{\footnotesize{\color{DarkBlue}{\tcp*{Update Param.}}}}

$w = w \odot \bm{M} + w^* \odot (1-\bm{M})$ \cref{eq:paramagg}
}
}
\end{algorithm}

\subsection{Discussion and Limitation}
\label{sec:discussion_and_limitation}

\noindent \textbf{Related Parameter Signal Investigation}. Generally speaking, parameter signals could be revealed in two aspects: magnitude {\cite{MagnitudePurning_NeurIPS15,LTH_ICLR19,Wanda_ICLR24}} and gradient {\cite{Fisher_1922,NatureGradient_arXiv13,EWC_PNAS17,Snip_ICLR19,StableCL_NeurIPS20,MovementPruning_NeurIPS20}}. The weight magnitude represents how much the parameter contributes to the prediction. The gradient reveals the information intensity during optimization. Thus, magnitude and gradient acts as parameter importance metrics to select target elements, which has incurred huge research interest in broad fields, such as network pruning {\cite{MagnitudePurning_NeurIPS15,LTH_ICLR19,PrAC_ICML21,PST_IJCAI22}}, domain generalization {{\cite{Fishr_ICML22,SAGM_CVPR23,ProGrad_ICCV23}}}, federated learning {\cite{FISH_NeurIPS21,FisherAveraging_NeurIPS22}}, and malicious defense {\cite{FedCPA_ICCV23,FTSAM_ICCV23,MMA_ICCV23,LeadFL_ICML23,SDFC_ECCV24}}. Existing explorations focus on training a network from scratch and face no requirement to preserve the previously learned knowledge, thus entangling the magnitude and gradient information to select the crucial elements for the target task. However, pre-trained \mllm{} (\mllmabbrv{}) models have inherent generalization knowledge, as evidenced by the capacity to execute diverse tasks without ﬁne-tuning {\cite{LLaVA_NeurIPS23,LLaVA15_CVPR24,VLMforVison_PAMI24}}. Thus, maintaining the generalization ability and enhancing the downstream specialization ability during the fine-tuning stage acts as a crucial task for \mllmabbrv{}. In our work, we utilize the pre-trained parameter magnitude ($\mathcal{I}$ in \cref{eq:magabsolute}) and optimizing parameter gradient ($\delta$ in \cref{eq:gradient}) to respectively reveal the parameter importance metrics for the generalization and specialization abilities. We further select relative downstream-kernel elements to balance the generalization and specialization ability during the fine-tuning process.

\begin{figure}[t]
\begin{center}
\includegraphics[width=\linewidth]{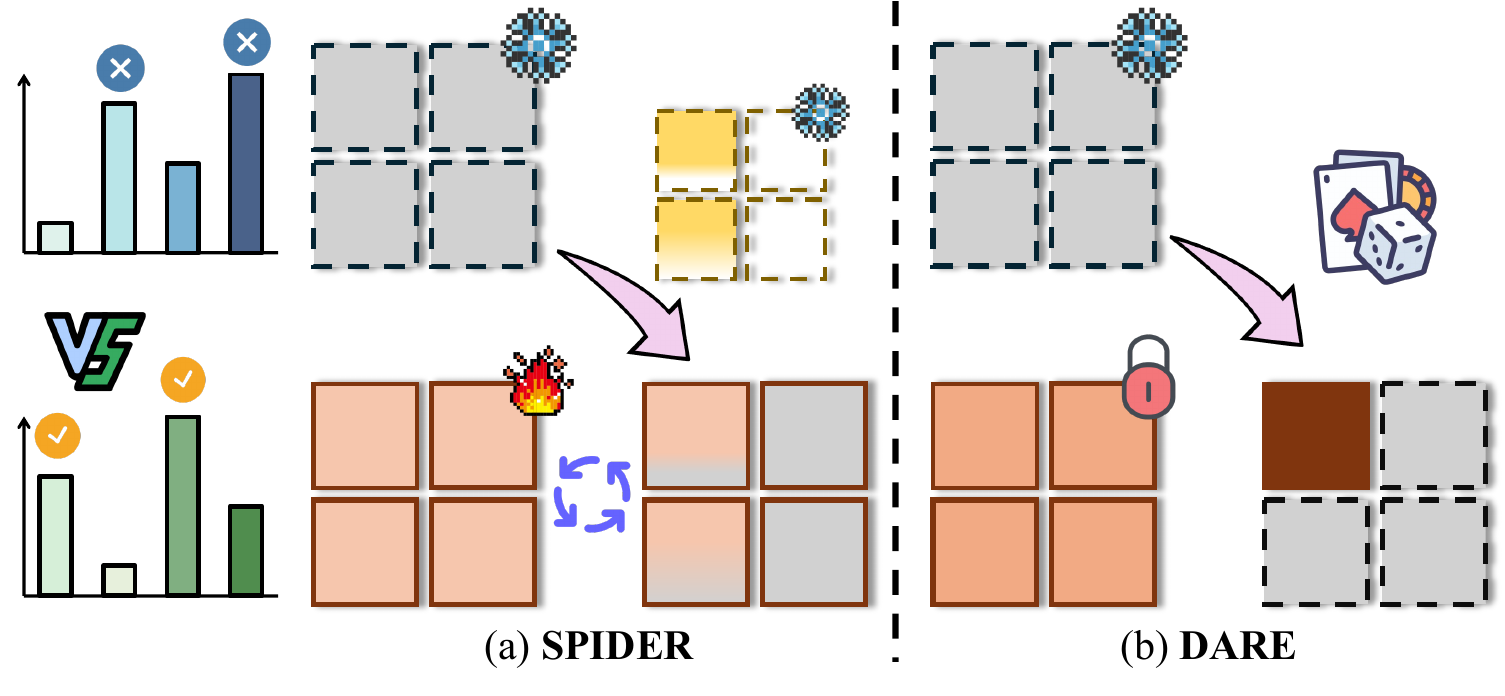}
\put(-230,63){\scriptsize{$\mathcal{I}[v]$ \cref{eq:magrescale}}} 
\put(-230,9){\scriptsize{$\mathcal{G}[v]$ \cref{eq:gradientweight}}}
\put(-127,104){\scriptsize{$\bm{M}[v]$}}
\put(-127,97){\scriptsize{\cref{eq:rescalemask}}}
\put(-127,53){\scriptsize{\cref{eq:paramagg}}}
\put(-30,6){\scriptsize{\cite{DARE_ICML24}}}
\end{center}
\vspace{-20pt}
\captionsetup{font=small}
\caption{\small\textbf{Conceptual Comparison.}  (a) \oursabbrv{}{} iteratively measures the parameter importance discrepancy to construct the update mask which protects generation and squeezes specialization information on the selected elements. (b) \dare{} combines the learned elements with the pre-trained and further rescale the candidate ones. \protect\includegraphics[scale=0.15,valign=b]{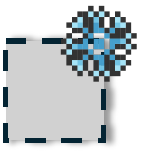} means frozen pre-trained elements. \protect\includegraphics[scale=0.15,valign=b]{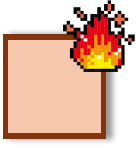} denotes current learning parameters. \protect\includegraphics[scale=0.15,valign=b]{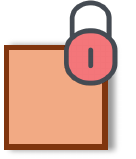} represents completed learned ones. }
\vspace{-15pt}
\label{fig:ConceptCompar}
\end{figure}

\noindent \textbf{Concept Difference}. Existing methods to mitigate \mllmabbrv{} forgetting, such as \dare{} {\cite{DARE_ICML24}} and \tailor{} {\cite{ModelTailor_ICML24}}, primarily focus on selectively updating and rescaling optimized parameters using random selection and the Hessian matrix {\cite{Fisher_1922,FISH_NeurIPS21,UnifiedDelta_arXiv24,SparseGPT_ICML23}}. However, these post-combination operations can conflict with optimization strategies that aim to adjust all trainable elements for downstream performance and is sensitive with the changing scale {\cite{TaskArithmetic_ICLR23,DARE_ICML24}}. Our approach evaluates parameter importance for both generalization and specialization objectives during the tuning stage. This enables us to selectively update parameters relevant to downstream tasks while preserving others, effectively performing an \textit{information extrusion} {\cite{LTH_ICLR19,UGS_ICML21,PrAC_ICML21,DLTH_ICLR22}} to reduce conflicts between pre-training and fine-tuning knowledge. We illustrate the concept difference in \cref{fig:ConceptCompar}.

\noindent \textbf{Limitation}. 
\oursabbrv{} leverages both previously pre-trained and current fine-tuning knowledge to select relevant downstream important elements and re-weight the candidate parameters, thereby ensuring generalization while pressing the fine-tuning optimization pathway. However, ours fails in certain circumstances. 
(i) We rank parameter importance based on the pre-trained weights and the current gradient matrix, which incurs additional memory usage. However, this increase is linear relative to the scale of learnable parameters, with a resource complexity of $\mathcal{O}(3 \times |\bm{M}|)$.
(ii) Our method assesses parameter importance from both pre-training and fine-tuning distributions, selecting task-relevant elements to balance generalization and specialization. When the downstream task closely aligns with the upstream distribution, fewer updates are needed, and minor distribution gaps introduce an acceptable level of generalization loss, managing the trade-off effectively.


\section{Experiments}
\label{sec:Experiments}

\subsection{Experimental Setup}
\label{sec:Setup}

\noindent \textbf{Architecture and Datasets}.
Adhering to the \mllm{} paradigm, we evaluate the effectiveness of our methods using two popular models as the foundations for our experiments: \llava{} {\cite{LLaVA_NeurIPS23}} and \vila{} \cite{VILA_CVPR24}. We categorize the datasets into two groups: pre-training (seen) and fine-tuning (unseen) datasets to respectively measure the generalization and specialization ability. The pre-training datasets consist of those used in the training process; accordingly, we assess the learned generalization ability on \okvqa{} \cite{OKVQA_CVPR19}, \textvqa{} \cite{TextVQA_CVPR19}, \gqa{} \cite{GQA_CVPR19}, and \ocrvqa{} {\cite{OCRVQA_ICDAR19}}. For fine-tuning tasks, we consider four downstream datasets: \flickrthirtyk{} {\cite{Flickr_TACL14}}, \cococap{} {\cite{COCO_ECCV14}}, \iconqa{} \cite{IconQA_NeurIPS21}, \scienceqa{} {\cite{ScienceQA_NeurIPS22}}\footnote{https://huggingface.co/datasets/BAAI/DataOptim}, which respectively associate with image caption and visual reasoning views. We follow {\cite{ReviEFT_arXiv22,PEFTMLLM_ACL24}} resource setting, and randomly sample $10k$ samples from the training set of each dataset.

\noindent \textbf{Counterparts}. We focus on exploring model-agnostic \mllmabbrv{} fine-tuning methods and mainly compare with the Regularization-based Optimization and Partial-based Updating solutions as follows:
\begin{fullitemize}
\item \fullft{} (\fullftabbr{}) \pub{arXiv'05} {\cite{CE_AOR05}}: Default optimize full  parameters towards the downstream task.
\item \ltworeg{} (\ltworegabbrv) \pub{PNAS'17} {\cite{EWC_PNAS17}}: Add an $\mathcal{L}_2$ regularization term with the regularization hyper-parameter, \ie, 1e-3, to the original loss function. Thus, it focuses on keeping the fine-tuning model closer to the pre-trained model, thereby mitigating forgetting.
\item \grafting{} \pub{ICML'23} {\cite{Grafting_ICML23}}:  Localize newly acquired skills
inside fine-tuned language models, which could be regarded as $\mathcal{L}_1$ regularization with the penalty weigh, \ie, 1e-6.
\item \halfft{} (\halfftabbr{}) \pub{arXiv'24} {\cite{HFT_arXiv24}}: Randomly update half of the parameter blocks within each transformer layer at each iteration while freezing the other elements.
\item  \dare{} \pub{ICML'24} {\cite{DARE_ICML24}}: Parameters from the fine-tuned model are randomly selected and re-scaled to maintain ability on generalization and specialization aspects.
\item  \tailor{} \pub{ICML'24} {\cite{ModelTailor_ICML24}}: Preserve pre-trained parameters while replacing a small ratio of fine-tuned parameters, \ie, 10 $\%$, based on the salience and sensitivity analysis.
\end{fullitemize}

\noindent \textbf{Implementation Details}. We follow the official codebase\footnote{https://github.com/haotian-liu/LLaVA}$^{,}$\footnote{https://github.com/NVlabs/VILA} to conduct the fine-tuning procedure. The learning rate $lr$ in \llava{} {\cite{LLaVA_NeurIPS23}} is $2e-4$ for \llmabbrv{} and $2e-5$ for the visual projector. For \vila{} {\cite{VILA_CVPR24}}, we uniformly set the learning rate to $1e-4$. The training epoch is set to $E=5$. The training batch size $B$ default set to 16. The fine-tuning block for \llmabbrv{} is the \textit{last} $L=2$ layers. All experiments are conducted on 8 NVIDIA 4090 GPUs, each with 24GB memory. Due to limited computation resources, we utilize the LLaVA-1.5-7B for \llava{} and VILA1.5-3B for \vila{}.

\noindent \textbf{Evaluation Metrics}. To evaluate the performance of \mllm{} (\mllmabbrv{}) in both generalization and specialization aspects, we consider two key metrics: Source Performance ($\mathcal{A}^\mathcal{S}$) and Target Performance ($\mathcal{A}^\mathcal{T}$). Let $\mathcal{U}=\{\mathcal{U}_i\}_{i=1}^{|\mathcal{U}|}$ represent the set of pre-training datasets and $\mathcal{T}$ denote the fine-tuning target dataset. Thus, we derive the following evalation metrics forms:
\begin{equation}\small
\setlength\abovedisplayskip{0pt} \setlength\belowdisplayskip{0pt}
\mathcal{A}^\mathcal{S} = \frac{1}{|\mathcal{U}|} \sum_i^{|\mathcal{U}|} \text{Acc.}(\mathcal{U}_i), \quad \mathcal{A}^\mathcal{T} = \text{Acc.}(\mathcal{T}).
\label{eq:gen_acc}
\end{equation}
Acc. denotes the accuracy metric. We use the CIDEr metric to evaluate performance on the \flickrthirtyk{} {\cite{Flickr_TACL14}} and \cococap{} {\cite{COCO_ECCV14}} datasets. For simplicity, we apply the same notation throughout. To evaluate effectiveness in mitigating catastrophic forgetting in \mllmabbrv{}, we use the H-Average metric ($\mathcal{H}$) and O-Average metric ($\mathcal{O}$) {\cite{ModelTailor_ICML24}}. The H-Average and O-Average metrics measure the harmonic and arithmetic mean of generalization ($\mathcal{A}^\mathcal{S}$) and specialization ($\mathcal{A}^\mathcal{T}$) performance as follows.
\begin{equation}\small
\setlength\abovedisplayskip{0pt} \setlength\belowdisplayskip{0pt}
\mathcal{H}= \frac{2 \times \mathcal{A}^\mathcal{S} \times \mathcal{A}^\mathcal{T}}{\mathcal{A}^\mathcal{S} + \mathcal{A}^\mathcal{T}}, \quad \mathcal{O}= \frac{\mathcal{A}^\mathcal{S} +  \mathcal{A}^\mathcal{T}}{2}.
\label{eq:harm_mean}
\end{equation}

\subsection{Diagnostic Analysis}
\label{sec:ablation}

We ablation on \flickrthirtyk{} and \iconqa{} for in-depth analysis.

\begin{table}[t]\small
\centering
{
\resizebox{\columnwidth}{!}{
\setlength\tabcolsep{5pt}
\renewcommand\arraystretch{1.1}
\begin{tabular}{c||cccIccc}
\hline \thickhline
\rowcolor{lightgray}
& \multicolumn{3}{cI}{\bm{\flickrthirtyk{}}} 
& \multicolumn{3}{c}{\bm{\cococap{}}} \\
\cline{2-7} 
\rowcolor{lightgray}
\multirow{-2}{*}{Metric}  
&  $\mathcal{A}^\mathcal{S}$ & $\mathcal{A}^\mathcal{T}$ & $\mathcal{H}$
&  $\mathcal{A}^\mathcal{S}$ & $\mathcal{A}^\mathcal{T}$ & $\mathcal{H}$
\\
\hline\hline
\fullftabbr{} 
& 47.04 & 66.68 & 55.16  
& 48.20 & 102.07 & 65.48 
\\ 
\hline
\multicolumn{7}{l}{$\bm{M}[v] = 1$, when $v$ satisfy [Metric]}
\\ 
\hline
$ \text{Rand}_{\gamma} $
& 51.68 &	70.15 &	59.52  
& 50.16 &	106.18& 68.13   
\\ 
$\mathcal{I}_{\gamma}$ \cref{eq:magrescale}
& 52.47 &	70.28 &	60.08 
& 51.07 &	106.10& 68.95  
\\ 
$\mathcal{G}_{\gamma}$ \cref{eq:gradientweight}
& 47.02 &	67.60 &	55.46 
& 47.99 &	105.33& 65.93  
\\ 
\rowcolor[HTML]{D7F6FF}
$\mathcal{G}[v] > \mathcal{I}[v]$  \cref{eq:binarymask}
& \textbf{52.68} &	\textbf{73.43} &	\textbf{61.35}  
& \textbf{51.70} &	\textbf{111.29}& \textbf{70.60}   
\end{tabular}}}
\vspace{-10pt}
\captionsetup{font=small}
\caption{\textbf{Ablation Analysis for Candidate Parameters Selection}. The $\text{Rand}_{\gamma} $ denotes randomly selects elements with $\gamma$ ratio. $\mathcal{I}_{\gamma}$ and $\mathcal{G}_{\gamma}$ respectively denotes choose $\gamma$ proportion  of elements via magnitude and gradient.  $\gamma$ is set as $50\%$.
Please see \cref{sec:ablation}.}
\label{tab:ablation_mask}
\vspace{-10pt}
\end{table}

\begin{table}[t]\small
\centering
{
\resizebox{\columnwidth}{!}{
\setlength\tabcolsep{5pt}
\renewcommand\arraystretch{1.1}
\begin{tabular}{cc||cccIccc}
\hline \thickhline
\rowcolor{lightgray} &
& \multicolumn{3}{cI}{\bm{\flickrthirtyk{}}} 
& \multicolumn{3}{c}{\bm{\cococap{}}} \\
\cline{3-8} 
\rowcolor{lightgray}
\multirow{-2}{*}{\firstabbrv{}} & \multirow{-2}{*}{\secondabbrv{}}  
&  $\mathcal{A}^\mathcal{S}$ & $\mathcal{A}^\mathcal{T}$ & $\mathcal{H}$
&  $\mathcal{A}^\mathcal{S}$ & $\mathcal{A}^\mathcal{T}$ & $\mathcal{H}$
\\
\hline\hline
\multicolumn{2}{c||}{\zeroshot{}} 
& 61.39 & 55.43 &58.26
& 61.39 & 107.64 & 78.19
\\ 
\hdashline
\multicolumn{2}{c||}{\fullftabbr{}} 
& 47.04 & 66.68 & 55.16  
& 48.20 & 102.07 & 65.48 
\\ 
\ding{51} & 
& 52.68 & 73.43 & 61.35    
& 51.70 & 111.29 & 70.60 
\\
\rowcolor[HTML]{D7F6FF}
\ding{51} & \ding{51}
& \textbf{55.40} & \textbf{83.49} & \textbf{66.61}  
& \textbf{55.94} & \textbf{122.74} & \textbf{76.86}   
\end{tabular}}}
\vspace{-10pt}
\captionsetup{font=small}
\caption{\textbf{Ablative Study of Key Modules} for \oursabbrv{}. Incorporate \textbf{sole} \first{} (\firstabbrv{}) can be regarded as \cref{eq:binarymask}. Considering \textbf{both} \firstabbrv{} and \secondabbrv{}, this is viewed as \cref{eq:rescalemask}. For a detailed discussion, please refer to \cref{sec:ablation}.}
\label{tab:ablation_module}
\vspace{-10pt}
\end{table}

\begin{figure}[t]
\centering
\includegraphics[width=\linewidth]{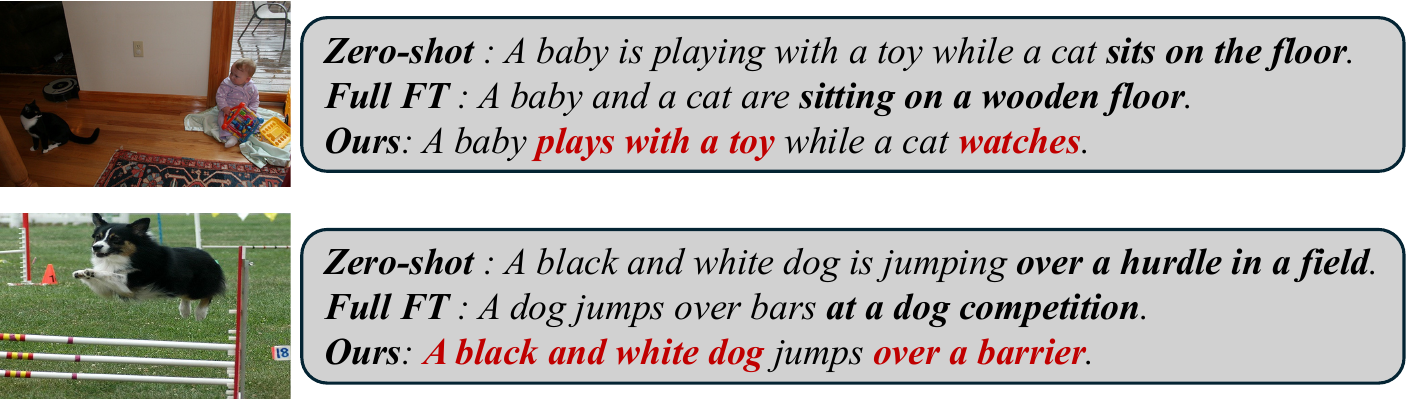}
\vspace{-15pt}
\captionsetup{font=small}
\caption{\textbf{Ablation Comparison on Response Output} on \flickrthirtyk{}. Text prompt is \textit{Write a short description for the image.} \fullftabbr{} better follow the instructions than \zeroshot, but \fullftabbr{} introduces hallucination (\eg, “at a dog competition”), while Zero-shot lacks task details. Please refer to \cref{sec:ablation}.}
\label{fig:example_compare}
\vspace{-15pt}
\end{figure}

\noindent \textbf{Candidate Parameter Selection Metrics}.
Selecting candidate parameters plays a crucial role in mitigating catastrophic forgetting of general knowledge while enhancing specialized behavior. In \cref{tab:ablation_mask}, we examine the impact of different mask updating strategies. Specifically, weight magnitude and gradient value serve as two effective metrics for assessing parameter importance relative to the current distribution. $\mathcal{I}_{\gamma}$ and $\mathcal{G}_{\gamma}$ denote the selection of $\gamma$ proportion of elements based on small pre-trained weight magnitude and large downstream gradient value, respectively. Three key observations emerge: {\ding{182}} Selecting partial parameters is an effective solution for balancing generalization and specialization. {\ding{183}} Solely considering the pre-training distribution ($\mathcal{I}_{\gamma}$) emphasizes generalization-related elements but significantly limits the downstream adaptation. {\ding{184}} Exclusively incorporating fine-tuning importance ($\mathcal{G}_{\gamma}$) undermines generalization and hinders specialization. As demonstrated in \cref{tab:ablation_mask}, we select relative downstream important elements ($\mathcal{G}[v] > \mathcal{I}[v]$ in \cref{eq:binarymask}), which effectively preserves generalization while ensuring specialization performance for \mllm{} (\mllmabbrv{}).




\noindent \textbf{Key Component Analysis}.
In {\cref{tab:ablation_module}}, we begin by validating the significance of our proposed components through their incremental integration. The first row displays the BASELINE result, representing a simple \fullft{} (\fullftabbr{}) approach using standard cross-entropy loss. As demonstrated, the combination of \first{} (\firstabbrv{}) and \second{} (\secondabbrv{}) yields the best performance in both generalization and specialization. This finding supports our motivation to evaluate parameter significance across pre-training and fine-tuning distributions, while selectively updating candidate parameters for the downstream task. Additionally, we plot the response outputs in \cref{fig:example_compare}, revealing that \zeroshot{} fails to adhere to the instruction style and lacks a detailed description. In contrast, naive \fullftabbr{} introduces hallucinations, likely due to the forgetting of generalized knowledge. Our method effectively achieves satisfying results.

\subsection{Comparison to State-of-the-Arts}
\label{sec:compSOTA}

\begin{table*}[t]\small
\centering
\scriptsize{
\resizebox{\linewidth}{!}{
\setlength\tabcolsep{3.pt}
\renewcommand\arraystretch{1.1}
\begin{tabular}{r||cccc|ccccIcccc|cccc}
\hline\thickhline
\rowcolor{gray!20}
 & 
\multicolumn{8}{cI}{\bm{\flickrthirtyk{}}} & 
\multicolumn{8}{c}{\bm{{\cococap{}}}}
\\
\cline{2-17} 
\rowcolor{gray!20}
\multirow{-2}{*}{Methods}  
& \okvqa{} & \textvqa{} & \ocrvqa{} & \gqa{} & $\mathcal{A}^\mathcal{S}$ & $\mathcal{A}^\mathcal{T}$ & $\mathcal{H}$ & $\mathcal{O}$
& \okvqa{} & \textvqa{} & \ocrvqa{} & \gqa{} & $\mathcal{A}^\mathcal{S}$ & $\mathcal{A}^\mathcal{T}$ & $\mathcal{H}$ & $\mathcal{O}$
\\
\hline\hline
\multicolumn{16}{l}{\textcolor{gray!60}{\textit{{Fine-Tune with \vila{}} architecture}}} 
\\ 
\zeroshot{} 
& 55.60 &	60.30 	&68.20 &	61.47 &	61.39 &	55.43 &	58.26 &	58.41 
& 55.60 &	60.30 	&68.20 &	61.47 &	61.39 &	107.64 &	78.19 &	84.52 
\\
\hdashline
\rowcolor{gray!10} \fullftabbr{} 
& 37.99 &	45.17	&53.85 &	51.14 &	47.04 &	66.68 &	55.16 &	56.86   
& 37.36 &	42.96 &	55.85 &	56.63 &	48.20 &	102.07 &	65.48 &	75.14 
\\

\ltworegabbrv{} 
& 34.59 &	25.89	&47.20 &	49.48 &	39.29 &	62.77 &	48.33 &	51.03 
& 33.98	&41.67	&52.55	&50.25	&44.61 &	99.84& 	61.67 	&72.23 
\\

\rowcolor{gray!10} \grafting{} 
& 35.66 &	31.60	&47.40 &	47.67 &	40.58 &	63.71 &	49.58 &	52.15   
& 33.12&	39.06	&52.60	&49.77 &	43.64 &	99.84 	&60.73 	&71.74 
\\

\halfftabbr{} 
& 44.15 &	48.71	&60.90 &	52.97 &	51.68 &	70.15 &	59.52 &	60.92 
& 41.41&	47.47&	57.80 &	53.94	&50.16 	&106.18 &	68.13 &	78.17 
\\

\rowcolor{gray!10} \dare{} 
& 38.38 &	39.69	&52.05 &	51.33 &	45.36 & 65.67 &	53.66 &	55.52  
& 36.73&	43.34&	56.5&	51.33	&46.98& 	100.70&	64.06 	&73.84 
\\

\tailor{} 
& 38.30 &	44.98&	53.35&	51.38&	47.00 &	65.00 & 54.56& 	56.00  
& 37.84 &	43.51& 	55.70& 	50.96& 	47.00 &	102.44&	64.44& 	74.72  
\\

\hline
\rowcolor[HTML]{D7F6FF}
\oursabbrv{} 
& 47.11 &	53.38	&65.55 &	55.57 &	\textbf{55.40} &	\textbf{83.49} &	\textbf{66.61}\redup{11.45} &	\textbf{69.45}\redup{12.59}   
&46.65	&54.94	& 65.55&	56.63	&\textbf{55.94} &	\textbf{122.74} &	\textbf{76.86}\redup{11.38} &	\textbf{89.34}\redup{14.20}  \\
\hline\hline
\multicolumn{16}{l}{\textcolor{gray!60}{\textit{{Fine-Tune with \llava{}} architecture}}} 
\\ 
\zeroshot{} 
& 58.00 &	58.25 &	66.20 &	61.93 &	61.10 &	25.31 &	35.79  &	43.20
& 58.00 &	58.25 &	66.20 &	61.93 &	61.10 &	110.52 &	78.69 &	85.81 
\\
\hdashline
\rowcolor{gray!10} \fullftabbr{}
& 45.59 &	47.09 &	57.65 &	56.94 &	51.82 &	61.58 &	56.28 &	56.70 
& 41.01 &	42.89 &	57.75 &	53.67 &	48.83 &	92.01 &	63.80 &	70.42 
\\
\halfftabbr{}
& 48.96 &	49.47 &	60.80 &	56.81 &	54.01 &	62.91 &	58.12  &	58.46
& 37.62 &	41.14 &	60.00 &	45.96 &	46.18 &	79.91 &	58.53 &	63.05 
\\
\rowcolor{gray!10} \dare{}
& 44.82 &	48.01 &	58.75 &	57.04 &	52.16 &	62.18 &	56.73 &	57.17 
& 39.60 &	39.45 &	56.00 &	51.50 &	46.64 &	90.82 &	61.63 &	68.73 
\\
\tailor{}
& 44.50 &	46.32&	59.00 &	57.14 &	51.74 &	61.27& 	56.10& 	56.51  
& 41.35 &	40.85&	58.45 &	54.87 &	48.88 &	90.94& 	63.58& 	69.91  
\\
\hline 
\rowcolor[HTML]{D7F6FF}
\oursabbrv{}
& 55.81 &	53.67 &	63.95 &	57.04   &	\textbf{57.62} &	\textbf{79.84} &	\textbf{66.93}\redup{10.65}  &	\textbf{68.73}\redup{12.03} 
& 49.50 &	45.33 &	65.00 &	58.14 &	\textbf{54.49} &	\textbf{114.74} &	\textbf{73.89}\redup{10.09} &	\textbf{84.62}\redup{14.20}
\end{tabular}}}
\vspace{-10pt}
\captionsetup{font=small}
\caption{{
\textbf{Comparison with the state-of-the-art \mllm{} (\mllmabbrv{}) Fine-Tuning Solutions} on the image caption task: \flickrthirtyk{} and \cococap{} datasets based on \vila{} and \llava{} architectures. We mark the Best in bold across different tuning methods. {\color{RedOrange}$\uparrow$}  means improved accuracy compared with Zero-shot.  Please refer to \cref{sec:compSOTA} for relative explanations. 
}}
\label{tab:compare_sota_image_capation}
\vspace{-10pt}
\end{table*}

\begin{table*}[t]\small
\centering
\scriptsize{
\resizebox{\linewidth}{!}{
\setlength\tabcolsep{3.pt}
\renewcommand\arraystretch{1.1}
\begin{tabular}{r||cccc|ccccIcccc|cccc}
\hline\thickhline
\rowcolor{gray!20}
 & 
\multicolumn{8}{cI}{\bm{\iconqa{}}} & 
\multicolumn{8}{c}{\bm{{\scienceqa{}}}}
\\
\cline{2-17} 
\rowcolor{gray!20}
\multirow{-2}{*}{Methods}  
& \okvqa{} & \textvqa{} & \ocrvqa{} & \gqa{} & $\mathcal{A}^\mathcal{S}$ & $\mathcal{A}^\mathcal{T}$ & $\mathcal{H}$ & $\mathcal{O}$
& \okvqa{} & \textvqa{} & \ocrvqa{} & \gqa{} & $\mathcal{A}^\mathcal{S}$ & $\mathcal{A}^\mathcal{T}$ & $\mathcal{H}$ & $\mathcal{O}$
\\
\hline\hline
\zeroshot{} 
& 55.60 &	60.30 &	68.20 &	61.47 &	61.39 &	19.93 &	30.09 &	40.66  
& 55.60 &	60.30 &	68.20 &	61.47 &	61.39 &	69.89 &	65.37 &	65.64  
\\
\hdashline
\rowcolor{gray!10} \fullftabbr{} 
& 34.51 &	38.02 &	46.10 & 	47.05 &	41.42 &	87.05 &	56.13 &	64.24 
& 47.15 &	50.88 &	57.20 & 	53.58 &	52.20 &	75.78 &	61.82 &	63.99 
\\

\ltworegabbrv{} 
& 21.69 &	25.89&	35.20 &	37.09 &	29.97 &	86.40 &	44.50 &	58.18 
& 43.65 &	48.13& 	51.80 &	50.42 &	48.50 &	76.40 &	59.33 &	62.45 
\\

\rowcolor{gray!10} \grafting{} 
& 22.66 &	31.60 &	40.25& 	37.84 &	33.09 &	87.18 &	47.97 &	60.13  
& 45.65 &	50.71 &	54.35& 	53.94 &	51.16 &	76.00 &	61.16 &	63.58 
\\

\halfftabbr{} 
& 43.36 &	48.71&	55.25 & 53.03& 	50.09 &	\textbf{88.19} &	63.89 &	69.14  
& 52.07 &	54.47&	60.55 &	57.52& 	56.15 &	76.77 &	64.86 &	66.46  
\\

\rowcolor{gray!10} \dare{} 
& 36.88	&   39.69&	45.15&	48.09&	42.45& 	88.11&	57.30& 	65.28  
& 47.61 &	50.39& 	57.55& 	55.08& 	52.66& 	\textbf{77.46}& 	62.69& 	65.06 
\\

\tailor{} 
& 37.99 &	41.42& 	48.10 &	47.57& 	43.77 &	88.16 &	58.50 &	65.97   
& 48.22 &	50.98& 	57.90 &	53.04& 	52.54 &	75.97 &	62.12 &	64.25 
\\

\hline
\rowcolor[HTML]{D7F6FF}
\oursabbrv{} 
& 48.34 &	53.38 &	63.35 &	56.12 &	\textbf{55.30} &	{84.07} &	\textbf{66.71}\redup{10.58} &	\textbf{69.68}\redup{5.44} 
& 54.13 &	57.33 &	65.60 &	60.07 &	\textbf{59.28} &	{75.97} &	\textbf{66.60}\redup{4.78} &	\textbf{67.63}\redup{3.64}  
\end{tabular}}}
\vspace{-10pt}
\captionsetup{font=small}
\caption{{
\textbf{Comparison with the state-of-the-art \mllm{} (\mllmabbrv{}) Fine-Tuning Solutions} on the the visual question answering task: \iconqa{}{} and \scienceqa{} datasets based on the \vila{} architecture. Please see details in \cref{sec:compSOTA}.
}}
\label{tab:compare_sota_vqa}
\vspace{-10pt}
\end{table*}

\begin{figure}[t]
\centering
\includegraphics[width=0.9\linewidth]{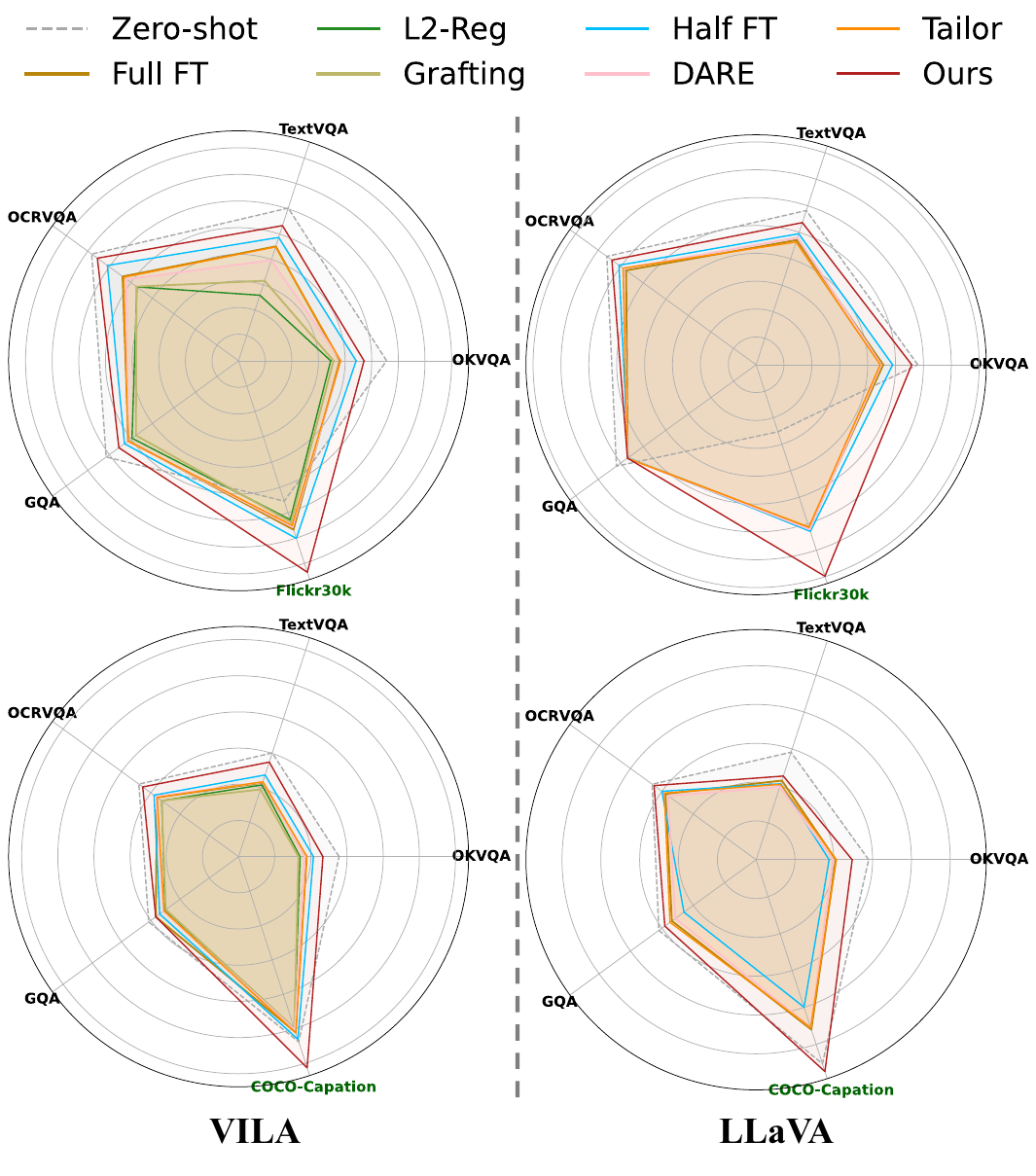}
\vspace{-10pt}
\captionsetup{font=small}
\caption{\textbf{Visualization Comparison}. Radar charts plots fine-tuning methods results across four pre-trained source datasets and target datasets, \ie,  {\color{DarkGreen}\flickrthirtyk{}} and {\color{DarkGreen}\cococap{}}. Our method achieves a better generalization and specialization trade-off.
} 
\label{fig:radar}
\vspace{-10pt}
\end{figure}

\noindent \textbf{Quantitative Results}.
We compare our \oursabbrv{} against related approaches on image-captioning (CAP) and visual question-answering (VQA) tasks. Due to the architectural complexity and task differences, we limit the VQA evaluation to \vila{}1.5-3B. As shown in \cref{tab:compare_sota_image_capation,tab:compare_sota_vqa}, several key observations can be made: The larger the task gap between the fine-tuning and pre-training distributions, the more severe the generalization-specialization trade-offs in the \mllmabbrv{} fine-tuning process. For example, we denote the $\alpha_\text{CAP}=\sum_{\mathcal{T}}\frac{\mathcal{H}_{\text{\textbf{\fullftabbr{}}},\mathcal{T}}-\mathcal{H}_{\text{\textbf{ZS}},\mathcal{T}}}{\mathcal{H}_{\text{\textbf{ZS}},\mathcal{T}}}$ ($\mathcal{T} \in \{\text{\flickrthirtyk{}},\text{\cococap{}}\})$. $\alpha_\text{VQA}=\sum_{\mathcal{T}}\frac{\mathcal{H}_{\text{\textbf{\fullftabbr{}}},\mathcal{T}}-\mathcal{H}_{\text{\textbf{ZS}},\mathcal{T}}}{\mathcal{H}_{\text{\textbf{ZS}},\mathcal{T}}}$ ($\mathcal{T} \in \{\text{\iconqa{}},\text{\scienceqa{}}\})$. Thus, $\alpha_\text{CAP}\!=\!-21.58 \% < \alpha_\text{VQA}\!=\!+81.11 \%$ for the \vila{} architecture. Notably, regularization-based optimization approaches typically offer limited performance improvements, as controlling parameter stiffness to regulate the extent of \llmabbrv{} updates remains a challenging task. Moreover, in partial-update methodologies, directly combining updated parameters with pre-trained ones often leads to performance fluctuation, largely due to the influence of delta parameters scale. In contrast, both \halfftabbr{} and our proposed method guide the \llmabbrv{} to fine-tune on selected parameters, demonstrating competitive performance across various experiments. Additionally, our approach selectively targets relatively important parameters for downstream tasks, yielding better performance compared to the random selection strategy employed in \halfftabbr{}. We further plot the radar visualization in  \cref{fig:radar} to highlight the performance advantages of ours compared to other approaches.

\begin{figure}[t]
\centering
\begin{subfigure}{0.50\columnwidth}
\includegraphics[width=\linewidth]{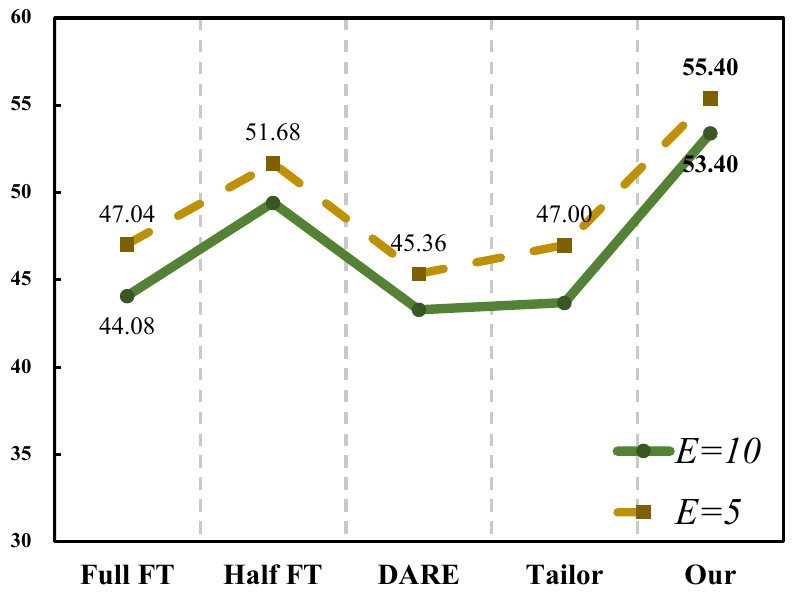}
\caption{\vila{} for  $\mathcal{A}^\mathcal{S}$}
\end{subfigure}
\hspace{-5pt}
\begin{subfigure}{0.50\columnwidth}
\includegraphics[width=\linewidth]{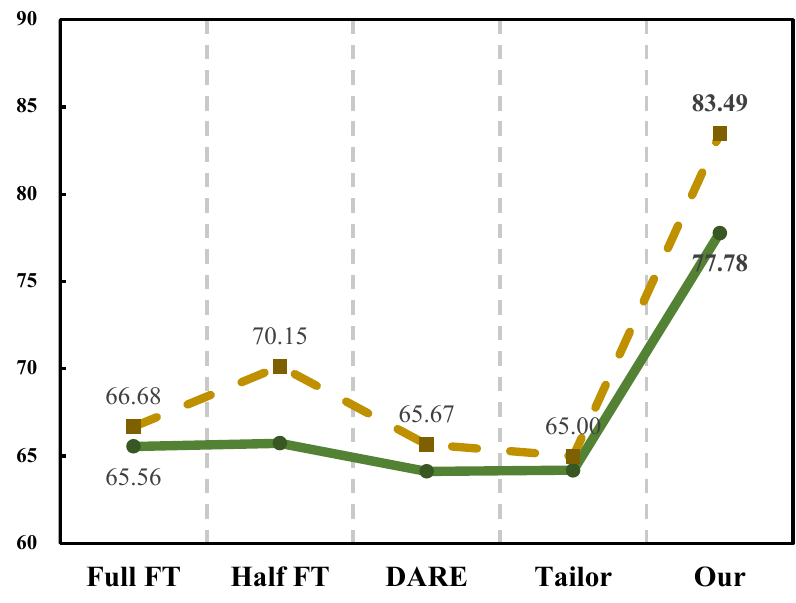}
\caption{\vila{} for  $\mathcal{A}^\mathcal{T}$}
\end{subfigure}
\hspace{-5pt}
\begin{subfigure}{0.50\columnwidth}
\includegraphics[width=\linewidth]{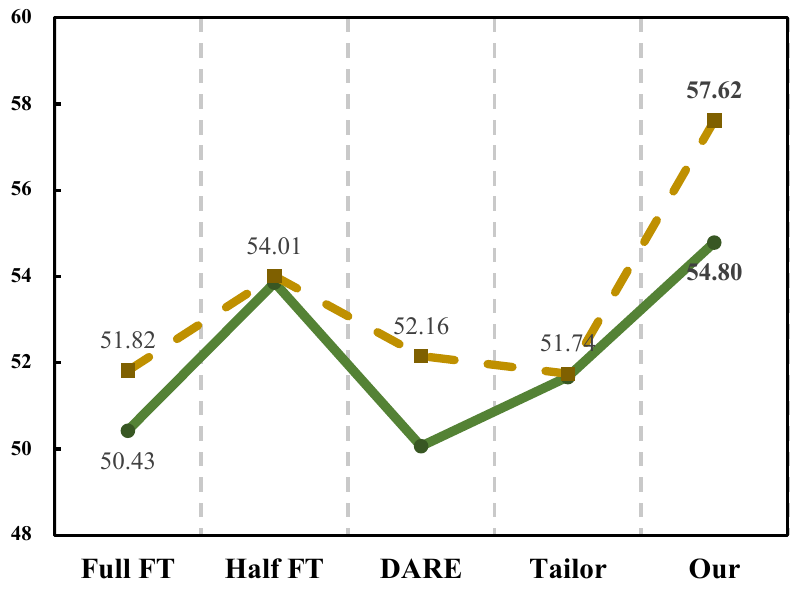}
\caption{\llava{} for  $\mathcal{A}^\mathcal{S}$}
\end{subfigure}
\hspace{-5pt}
\begin{subfigure}{0.50\columnwidth}
\includegraphics[width=\linewidth]{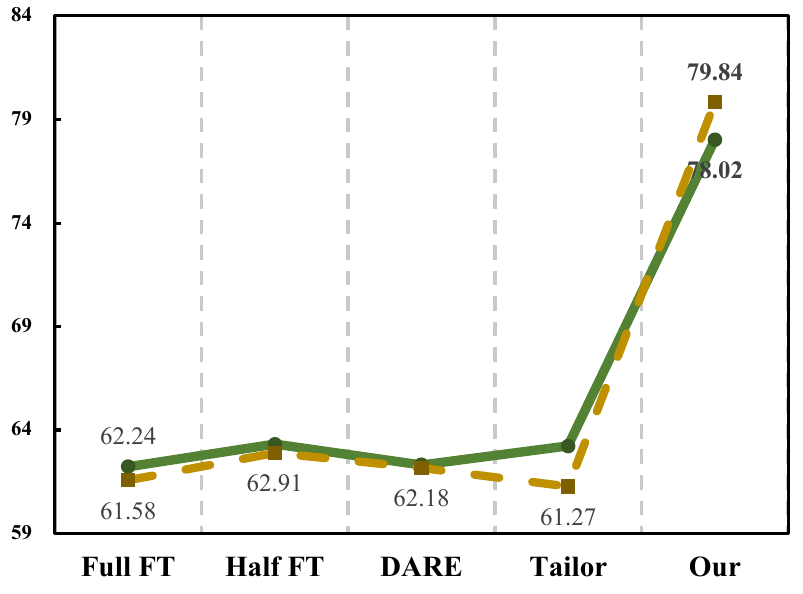}
\caption{\llava{} for  $\mathcal{A}^\mathcal{T}$}
\end{subfigure}
\vspace{-20pt}
\captionsetup{font=small}   
\caption{\small{
\textbf{Comparison on  Large Fine-Tuning Epochs }$E$ from ({\color{DarkYellow}5 rounds} to {\color{DarkGreen}10 rounds}) on  \flickrthirtyk{}. Refer to \cref{sec:compSOTA} for details.
}}
\label{fig:ablation_epoch}
\vspace{-15pt}
\end{figure}

\noindent \textbf{Performance on More Tuning Epochs} $E$. We investigate the impact of extending fine-tuning epochs $E$ from 5 to 10 rounds, as shown in \cref{fig:ablation_epoch}. The results highlight several key findings: (i) Extending fine-tuning epochs intensifies the pre-training knowledge forgetting phenomenon across different architecture scales. 
(ii) Smaller architectures, such as \vila{}-1.5-3B, encounter more severe parameter conflicts, where a decline in generalization ability results in degraded specialization performance. For instance, when extending the number of epochs from 5 to 10, $\mathcal{A}^\mathcal{S}\!: \! 47.04 \! \rightarrow \! 44.08$ and $\mathcal{A}^\mathcal{T}\!: \!66.68 \! \rightarrow  \! 65.56$ for \fullft{}.
(iii) Larger architectures, such as \llava{}-1.5-7B, which possess higher parameter redundancy, maintain more stable specialization ability despite extended tuning epochs. Specifically, a lightly $\mathcal{A}^\mathcal{T}$ score increases: $+0.66$ for \fullft{}.
(iv) Our method achieves robust performance across various architectures and training duration.

\begin{figure}[t]
\centering
\begin{subfigure}{0.50\columnwidth}
\includegraphics[width=\linewidth]{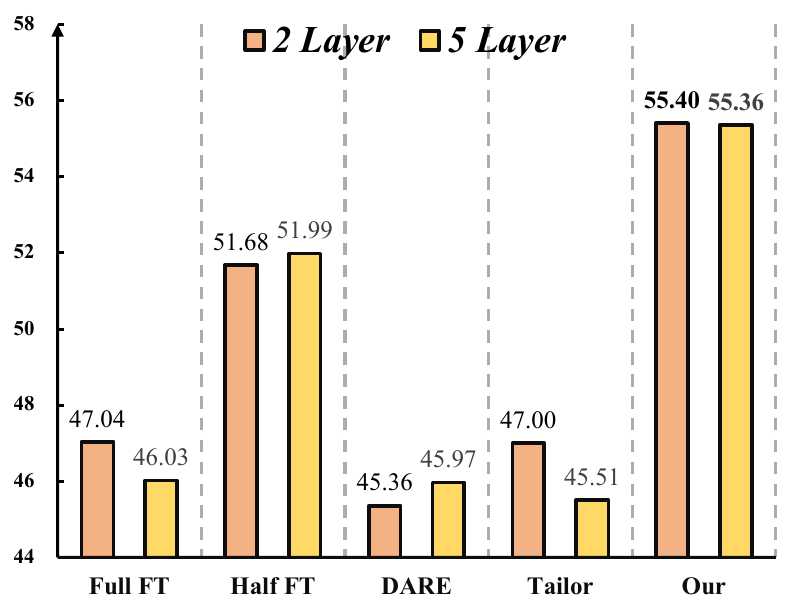}
\caption{Generalization ($\mathcal{A}^\mathcal{S}$)}
\end{subfigure}
\hspace{-5pt}
\begin{subfigure}{0.50\columnwidth}
\includegraphics[width=\linewidth]{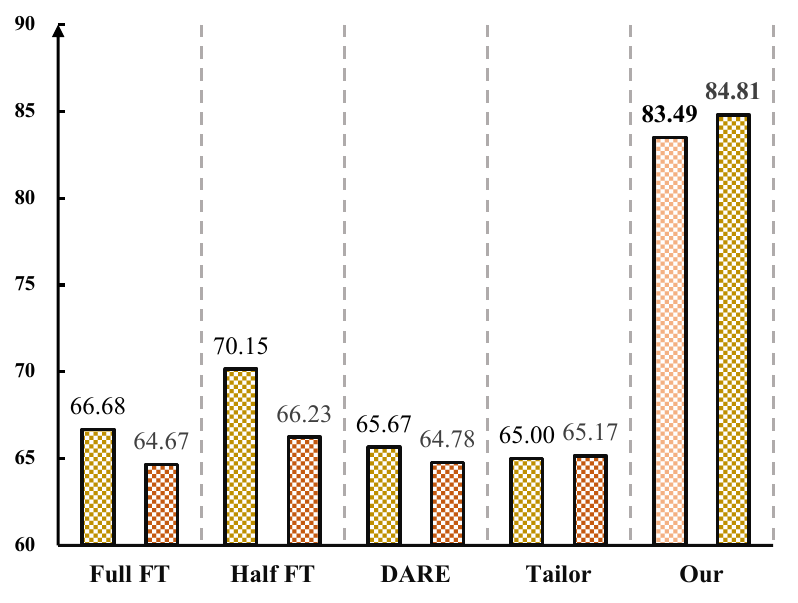}
\caption{Specialization ($\mathcal{A}^\mathcal{T}$)}
\end{subfigure}
\vspace{-20pt}
\captionsetup{font=small}   
\caption{\small{
\textbf{Comparison on More Fine-Tuning Layer} $L$ on \flickrthirtyk{} dataset with the \vila{} architecture. See \cref{sec:compSOTA}.
}}
\label{fig:ablation_layer}
\vspace{-15pt}
\end{figure}

\noindent \textbf{Performance on More \llmabbrv{} Tuning Layer} $L$. We evaluate the effect of tuning block layers $L$ from 2 to 5, as shown in \cref{fig:ablation_layer}. Existing methods show a slight improvement in general performance but significantly reduce target domain performance. \oursabbrv{} effectively maintains both generalization and specialization across various tuning layers. 

\begin{figure}[t]
\centering
\begin{subfigure}{0.50\columnwidth}
\includegraphics[width=\linewidth]{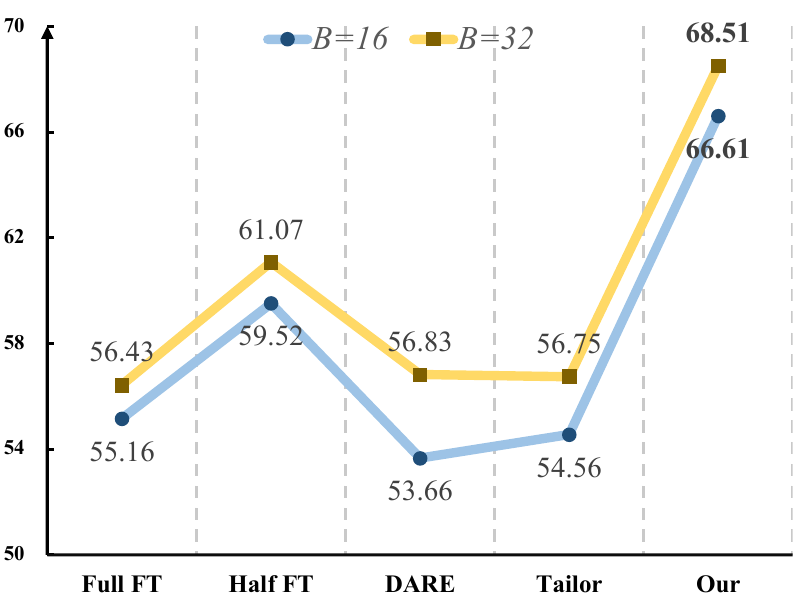}
\caption{\vila{} for $\mathcal{H}$}
\end{subfigure}
\hspace{-5pt}
\begin{subfigure}{0.50\columnwidth}
\includegraphics[width=\linewidth]{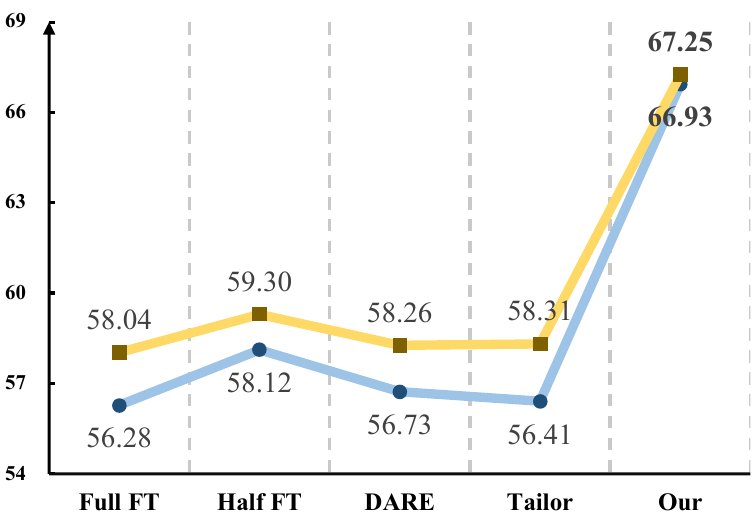}
\caption{\llava{} for $\mathcal{H}$}
\end{subfigure}
\vspace{-20pt}
\captionsetup{font=small}   
\caption{\small{
\textbf{Comparison on Large Training Batch Size} $B$ on \flickrthirtyk{} dataset with  \vila{} and \llava{}. Refer to \cref{sec:compSOTA}.
}}
\label{fig:ablation_batch}
\vspace{-15pt}
\end{figure}

\noindent \textbf{Performance on Large Training Batch} $B$. We further conduct the experiments on the large training batch $B$: 32 in \cref{fig:ablation_batch}. The results show that setting a higher training batch benefits both generalization and specialization ability across different counterparts. Notably, our method \oursabbrv{} constantly achieves the best performance results.

\section{Conclusion}
In conclusion, we address the catastrophic forgetting in fine-tuning \mllm{} (\mllmabbrv{}). We introduce \ours{} (\oursabbrv{} \protect\includegraphics[scale=0.1,valign=c]{Figure/SPIDER.pdf}), a novel approach to assess parameter importance for both generalization and specialization, focusing on identifying downstream-important elements and performing critical-aware updates on selected parameters. Our method enjoys third advantages: First, {\textit{No Architecture Dependency}}: \oursabbrv{} functions without specific model architecture, which presents high transferability across different architectures. Second, {\textit{No Fine-tuning Pattern Conflict}}: we conduct partial parameter updates, maintaining compatibility with various optimization functions. Third, {\textit{No Hyper-Parameter Configuration}}: leveraging parameter importance discrepancies requires no additional hyper-parameters, enhancing fine-tuning effectiveness. \oursabbrv{} has been validated on fruitful scenarios, highlighting the potential for broader applications.

{
\small
\bibliographystyle{ieeenat_fullname}
\bibliography{reference}
}

\end{document}